%% file: asme2ej.tex
\documentclass[subscriptcorrection,upint,varvw,barcolor=Goldenrod3,mathalfa=cal=euler,balance,hyphenate,english,pdf-a]{asmejour} %

\hypersetup{%
	pdfauthor={Mariana Pinto},                       		   	
	pdftitle={Data and Knowledge for Overtaking Scenarios in Autonomous Driving},                  	
	pdfkeywords={ASME journal paper, LaTeX template, BibTeX style, asmejour class},
	pdfsubject = {Automated driving survey plus dataset with focus on overtaking},			
}

\usepackage{epsfig} 
\usepackage{array}
\usepackage{longtable} 
\usepackage{multicol}
\usepackage{booktabs}
\usepackage{graphicx}
\usepackage{multirow}
\usepackage{diagbox} 
\usepackage{array}
\usepackage{makecell}
\usepackage{subcaption}
\usepackage[normalem]{ulem}
\usepackage{tabularx}
\usepackage{color, colortbl}
\definecolor{myGray}{rgb}{222,219,219}
\definecolor{green}{cmyk}{0.1,0,0.15,0.06}
\definecolor{blue}{cmyk}{0.31,0.15,0.0,0.06}
\usepackage{float}
\usepackage[export]{adjustbox} 
\usepackage{capt-of} 
\usepackage{multicol} 
\usepackage{balance}
\newsavebox{\measurebox}
 
\usepackage{hyperref}
\usepackage[nolist]{acronym}

\newcommand{\ts}{\textsuperscript} 

\definecolor{codeblue}{rgb}{0.01, 0.28, 1.0}
\definecolor{codegreen}{rgb}{0.01, 0.28, 1.0}
\definecolor{codegray}{rgb}{0.5,0.5,0.5}
\definecolor{codered}{rgb}{0.82, 0.1, 0.26}
\definecolor{backcolour}{rgb}{0.95, 0.95, 0.96}
\definecolor{plot_blue}{HTML}{000066}
\definecolor{plot_green}{HTML}{006600}

\JourName{Auton. Vehicles and Systems}

\begin{document}



\SetAuthorBlock{Mariana Pinto\CorrespondingAuthor}{Department of Computer Science\\
	Faculty of Sciences\\
	University of Porto\\
	Porto, Portugal, 4169-007\\
    Email: marfpp19@gmail.com} 


\SetAuthorBlock{In\^es Dutra}{%
Department of Computer Science\\
	Faculty of Sciences and CINTESIS@RISE\\
	University of Porto\\
	Porto, Portugal, 4169-007\\
    Email: ines@dcc.fc.up.pt
}

\SetAuthorBlock{Joaquim Fonseca}{
    Bosch Car Multimédia Portugal, S.A.\\
    Braga, Portugal, 4701-970\\
    Email: Joaquim.Fonseca2@pt.bosch.com
}

\title{Data and Knowledge for Overtaking Scenarios in Autonomous Driving}


\begin{abstract}
{\it Autonomous driving has become one of the most popular research topics within Artificial Intelligence. An autonomous vehicle is understood as a system that combines perception, decision-making, planning, and control. All of those tasks require that the vehicle collects surrounding data in order to make a good decision and action. In particular, the overtaking manoeuvre is one of the most critical actions of driving. The process involves lane changes, acceleration and deceleration actions, and estimation of the speed and distance of the vehicle in front or in the lane in which it is moving. Despite the amount of work available in the literature, just a few handle overtaking manouevres and, because overtaking can be risky, no real-world dataset is available. This work contributes in this area by presenting a new synthetic dataset whose focus is the overtaking manouevre. We start by performing a thorough review of the state of the art in autonomous driving and then explore the main datasets found in the literature (public and private, synthetic and real), highlighting their limitations, and suggesting a new set of features whose focus is the overtaking manouevre.


}
\end{abstract}

\maketitle

\begin{acronym}
\acro{ADAS}{Advanced Driver Assistance Systems}
\acro{AI}{Artificial Intelligence}
\acro{AUC}{Area Under the Curve}
\acro{AV}{Autonomous Vehicle}
\acro{CNN}{Convolutional Neural Networks}
\acro{DNN}{Deep Neural Network}
\acro{ER}{Entity Relationship}
\acro{GAN}{Generative Adversarial Network}
\acro{ILP}{Inductive Logic Programming}
\acro{ITS}{Intelligent Transport Systems}
\acro{KB}{Knowledge Base}
\acro{KR}{Knowledge Representation}
\acro{KRR}{Knowledge Representation and Reasoning}
\acro{LiDAR}{Light Detection and Ranging}
\acro{NHTSA}{National Highway Traffic Safety Administration}
\acro{RNN}{Recurrent Neural Network}
\acro{ROC}{Receiver Operating Characteristic}
\acro{ROS}{Robot Operating System}
\acro{SAE}{Society of Automotive Engineers}
\acro{SBS}{Sequential Backward Selection}
\acro{SMOTE}{Synthetic Minority Oversampling TEchnique}
\acro{SVM}{Support Vector Machines}
\acro{YAP}{Yet Another Prolog}

\end{acronym}



\section{Introduction}

Autonomous driving is a widely discussed topic nowadays. There are several papers in the literature that overview the myriad possibilities and problems that drive the impact in this field of research. 
Autonomous driving as we know it today is based on comfort, safety and promises to revolutionize transportation services. According to the World Health Organization, in the UN Global Road Safety Week 2021 report, every year more than 1.35 million people are killed in road accidents worldwide, which means almost 700 deaths on the roads every day~\cite{seguranca_rodoviaria_2021_2021}. In the recent past (2007–2014), of the total reported crashes, nearly 37\% occurred on national highways, mainly two-lane two-way with mixed traffic environments~\cite{crashes_report}.
In order to decrease these numbers, many countries develop their road safety plans based on the ``Vision Zero'' system. The term was conceived in Sweden in 1997, and it can be summarized in one sentence: No loss of life is acceptable~\cite{tingvall1999vision}. 

Despite the advantages that autonomous driving can bring, it still faces numerous challenges. It is therefore important to understand how these challenges can be mitigated and how much we can rely on the data we have available today.

In the next sections, we discuss about the current status of autonomous driving. We then move to discuss about the needed inputs for decision-making in the context of overtaking. We propose a set of features found relevant to make the decision and, finally, close with conclusions and perspectives of future work.

\section{Current State of Autonomous Driving}
The current worldwide panorama of competitors in the autonomous driving market is vast and complete, especially involving companies such as: Waymo/Google/Alphabet, Cruise, Mobileye, Apollo, Baidu, Bosch, Voyage, Aurora, Wayve, Tesla, Apple, Cisco, Aptiv, Alibaba, Drive.ai, Intel Corporation, Daimler/Mercedes-Benz, Audi, BMW, Ford Motors Corporation, General Motors Company, Honda, Hyundai Motor Company, Huawei, NIO, SenseTime, Uber ATG, Zoox, Samsung, Qualcomm, Jaguar Land Rover, PSA Group, Toyota Motor Corporation, AEye, Magna, AutoX, lyft, navya, Valeo, Continental, Denso, HERE, and others.
There are many advances in this area thanks to the investments and several studies carried out by all these companies using different methodologies, strategies, and experiments. Some recent examples are: Amazon announced an investment of 700 million Euros in Rivian, a direct competitor of Tesla~\cite{amazon_rivian}; Audi established a partnership with NVIDIA~\cite{nvidia_audi}; BMW announced a partnership with Aurora and Apollo~\cite{apollo_bmw}; among others.

According to a McKinsey\&Company's analysis, up to 15\% of new cars sold in 2030 could be fully autonomous meaning that possibly 90\% of the accidents can be potentially decreased when autonomous vehicles are deployed~\cite{mckinsey}\cite{Bertoncello2021Jun}. Nowadays, although there are no completely autonomous cars, they present many features that can help and alert the driver, thus enhancing the driving experience. 
While some systems already allow the driver to take their hands off the wheel, in certain situations, none yet allow drivers to safely take their eyes off the road. The systems that are currently implemented are described in Table~\ref{tab:automated_systems} placed in Appendix~\ref{appendixA}. 

There are still many challenges before autonomous driving can gain the trust of road users and achieve the desired levels of safety and comfort. Liu et al.~\cite{liu2020computing} conclude that while humans maintain an advantage in perceiving and sensing the environment, a combination of sensors can do a better job, especially in adverse weather or low lighting conditions. The following challenges are also mentioned: interaction of road users with machines; multisensory data synchronization – handle a variety of data sources and synchronize them; energy consumption – it's a challenge due to the amount of sensors and computing devices implemented in the vehicle; data protection – the data has to be protected in order not to be vulnerable to cyberattacks; scarcity of labelled data – companies holding this data are the ones that are one step ahead in the race to autonomous driving; accumulate and hold massive amounts of driving data for application of Machine/Deep Learning technologies – only then can knowledge and algorithms be scaled with the necessary safety~\cite{liu2020computing}.
Due to all the challenges and issues that autonomous driving cannot yet address, user uncertainty in relying on this technology is still significant. The latest AAA annual survey on automated vehicles shows that just 14\% of drivers would trust riding in a self-driving car. 54\% claim to be afraid to drive an autonomous vehicle and 32\% are not sure about the subject~\cite{aaa}.

Many studies are conducted in order to solve the issues described above. Some studies focus on understanding driver behaviour and its influence on the road environment. According to ~\cite{Basu_Yang_Hungerman_Singhal_Dragan_2017}, over 80\% of users prefer the style that they think is their own, but very often they were incorrect in identifying their own style.
In the literature we can also find algorithms that adapt the autonomous car behaviour depending on what other drivers are doing~\cite{8593682} or a dynamic system between an autonomous car and a human driver~\cite{Sadigh-RSS-16}. In the latter, it is proved that autonomous car’s actions will actually affect what other cars will do in response. The authors model these consequences by approximating the human as an optimal planner, with a reward function that they acquire through Inverse Reinforcement Learning which is the problem of inferring the reward function of an agent, given its policy or observed behaviour~\cite{ARORA2021103500}.
Due to the danger present and the complexity of the driving scenarios, one of the solutions is to create middle-level input and output representations which allow easy mixing of real and simulated data~\cite{Bansal_Krizhevsky_Ogale_2019}. In fact, the authors train a policy for autonomous driving via imitation learning and augment the imitation loss with additional losses that penalize undesirable events and encourage progress, using the ChauffeurNet model (a \ac{RNN}) to simulate complex traffic situations. An excellent proposal for systematic review of public driving datasets and toolsets for autonomous driving virtual test is presented by Ji {\it et al.}~\cite{9564428}. Some of the suggestions are used in this paper.

\section{Inputs for Decision-Making} 
Autonomous vehicles are systems that combine 4 modules: perception, decision-making, planning, and control. The decision-making module is said to be the brain of self-driving cars. This procedure involves obtaining information from the perception module and then making corresponding decisions for the planning and the control modules~\cite{6318436}. Due to the variety of situations that can occur while driving, the challenges of decision-making are many and complex. Proof of this is the critical difference between level 2 and levels 3/4/5 on the Society of Automotive Engineers (SAE) scale of autonomy levels. The difference consists in the vehicle's ability to make decisions such as overtaking or changing lane on its own~\cite{autonomy}.

The autonomous vehicle's decision-making mechanism consists of the transition from perceiving world surrounding the vehicle to the motion planning system. 
In general, environmental cues and the status of the ego vehicle\footnote{Term used to designate the automated vehicle that contains the sensors that perceive the environment around it.}, the one that tries to overtake, are inputs to a decision-making system, and the outputs are strategies that include behaviours and control commands that are fed into the motion planning system. Sensors placed in automobiles capture data from the surrounding environment, which is subsequently processed to provide perceptual outputs. The ego vehicle state is a representation of its current location and movement. Information at the lane level can also be used to enhance the perception system. Outputs, in turn, are represented by behaviors such as overtaking, lane changing, among others, and control commands including acceleration, angular speed etc~\cite{inputs_decision}.

\subsection{Perception}
Perception is the base and the vital part of all autonomous driving processes~\cite{VANBRUMMELEN2018384}. It mimics the human sensory system by analysing and understanding the surrounding environment. The outcomes of perception influence all the other areas. For instance, like it happened in 2018 when an Uber self-driving car killed a pedestrian holding a bike while crossing an urban street in the night. According to a report by the \ac{NHTSA}, Uber’s sensors were able to detect the pedestrian, but only identified them as an ``unknown object''~\cite{Uber_accident}. This accident exposes the need for more robust perception algorithms that are able to deal with multiple driving situations. Nowadays, perceptions algorithms are very complex and present good results in ``ideal'' street conditions, \textit{i.e.}, streets with excellent road markings, flat and weakly curved road and perfect weather conditions. However, real-world situations are unpredictable, complex, and even dangerous. According to the literature, there are still many challenges in this knowledge area to obtain an efficient and trusted perception, namely perception in adverse weather and poor light conditions, perception in complex environments and high velocities, real-time response and explainability, security metrics in case of sensor failure and others.

Chaabani et al. focused their study on the problems caused by the lack of visibility in foggy conditions. They present a neural network approach for estimating visibility distances and evaluating the solution using a diverse set of images under various fog density scenarios.The results suggested better performance compared to classical methods such as Linear Regression or Support Vector Machine~\cite{CHAABANI2017466}. Another highly respected work is the one conducted by~\cite{736004}, where the authors used a wavelet transform (wave-like oscillation with an amplitude that begins at zero, increases or decreases, and then returns to zero one or more times) for estimating the distance of visibility in fog conditions. The physical and probabilistic models allowed them to detect the highest edge of the image with a contrast over 5\%. 

The perception of the environment can be split into 3 steps: collecting, filtering and processing the raw data obtained from the sensors with information that could come from the communications, the infrastructure and/or other vehicles. The perception consists of estimating the attributes of the 5 key components of the road scene. These 5 key components are:
\begin{enumerate}
    \item \textbf{obstacles}: objects that can be found in the surrounding area of an ego vehicle. The objects can be static (e.g. a hole) or dynamic (other vehicles).
    \item \textbf{road}: attributes that define the road surface (lanes, road markings, road states and types).
    \item \textbf{ego vehicle}: position and dynamic state.
    \item \textbf{environment}: weather conditions and other road-side components like vertical road signs or off-road objects.
    \item \textbf{driver}: current state of the driver (vigilance, drowsiness, etc.).
\end{enumerate}

For driving automation, only the first four components are required. It is necessary to detect obstacles to avoid collisions and to have access to the entire road area surrounding the ego vehicle. By defining the road surface, we can obtain the number of lanes, identify which lane the ego vehicle is in, and recognize the road markings~\cite{VANBRUMMELEN2018384}. 
The study of the impact of adverse weather conditions for autonomous driving has been getting more attention recently. One example is the study conducted by~\cite{weather_conditions} that presents the effect of weather on different sensors used in autonomous vehicles and shows that the detection range of radar can be reduced by up to 45\% under heavy precipitation conditions.
One of the most complicated tasks for perception is the detection and classification of occluded objects. Overtaking, for example, can be a challenging task when the view of the road is obstructed by a truck that makes it impossible to see through it. \cite{vanet} and \cite{dynamic_seethroughs} show different approaches to the topic. In the first, the authors propose an STS (See-Through System) system that is based on VANET (Ad hoc Networks) and video streaming technology. This system increases the driver's visibility and supports his decision to overtake in challenging situations. In the second, a 2D projective invariant is used to capture information about occluded objects (which may be moving). There have been other more recent studies with a game theory-based approach such as the study conducted by~\cite{safe_occlusion} which introduces a novel analytical approach that posits safe trajectory planning under occlusions as a hybrid zero-sum dynamic game between the autonomous vehicle (evader) and an initially hidden traffic participant (pursuer). The analysis produces optimal strategies for both players that avoid collisions and support decision-making. 

Autonomous driving is receiving increasing attention, with more and more resources becoming available to enable safe, reliable, and efficient automated mobility in complex and uncontrolled environments. Signal processing is a critical component of autonomous driving. Some required technologies include affordable sensing platforms that can acquire data under varying environmental conditions, reliable simultaneous localization and mapping, machine learning that can effectively handle real-world conditions and unforeseen events, complex algorithms to bring more effective classification and decision-making, efficient real-time performance, resilient and robust platforms that can withstand breaches and adversarial attacks, and end-to-end system integration of detection, signal processing, and control. 
The data processing is structured in five levels. The first level corresponds to the raw data collected by the sensors (laser impacts, RADAR frames, LiDAR point clouds, distances, speeds, images, accelerations, angles). The second level focuses on filtering, spatial and temporal alignment, modelling of inaccuracy, uncertainty, and reliability. The third level corresponds to clustering, feature extraction, object detection and modelling. In the next level, more detailed information is obtained, such as the shape, colour and positioning of the object. At the last level, the interactions between objects are treated in order to build a more synthetic and enriched representation. The temporal relationships between objects make it possible to identify some behaviours and predict trajectories. The results of the last level can be used as inputs for the decision level (decision-making, risk assessment or trajectory generation)~\cite{VANBRUMMELEN2018384}.

The detection of obstacles of various shapes, sizes, and orientations is still a challenge due to the lack of information in the literature and to the number of labelled and tagged datasets which is still scarce.
The system for recognizing the environment around the vehicle must be done robustly, accurately and at a 360° angle by a combination of sensors. Several companies like Audi, Bosch Group, Uber, and Google/Waymo consider the LiDAR as a pillar sensor during the perception phase, while others like Tesla and Nissan do not consider it essential possibly because of cost-performance trade-off since the cost of placing a single LiDAR device on a car is around \$10,000~\cite{yan2016can}\cite{sharabok_2020}.
These values, however, are dependent on the sensor's robustness and range from \$1,000 to \$23,000~\cite{lidar_cost_1}~\cite{lidar_cost_2}.
Instead of LiDAR, they prefer to use radar sensors and cameras. On the other hand, the radar sensor has its value recognized for being particularly useful in environments with adverse weather conditions (fog, rain, snow). However, to obtain 360º coverage would require many radar sensors, as demonstrated in Aptiv's nuScenes dataset which uses 5 radar units in its setup and still fails to achieve 360º mapping at close range~\cite{9156412}.

\subsection{Datasets}
To obtain robust and accurate perception algorithms, hundreds of millions of data are required. To train the algorithms, the stock of training data must be labeled (``ground truth'') to ensure high training accuracy~\cite{Kalra_miles}.
Ground truth refers to the actual values of the training set’s classification, it is the manual verification that the virtual world corresponds to the real human-defined measurements. 
The road environment as we know is quite unpredictable and some scenarios are too dangerous to be staged in the physical world (eg. a child behind the car during parking). Nevertheless, some companies like Uber, Google or Tesla usually test their vehicles in real traffic conditions. This brings disadvantages in that it is very difficult to reproduce dangerous and imprecise scenarios and makes it impossible to get the ``ground truth'' of obstacles, pedestrians, and other vehicles involved in the test~\cite{Huang2016Nov}. For this reason, an alternative is to train and validate driving strategies in simulation with synthetic data. Table~\ref{tab:datasets_comp} placed in Appendix~\ref{appendixB}, shows a comparison of the features of the main datasets for autonomous driving. 

According to the Gartner ‘Predict’, published in the Wall Street Journal by analyst Svetlana Sicular, ``By 2024, 60\% of the data used for the development of AI and analytics projects will be synthetically generated''~\cite{svetlana}.
Some famous datasets are public and available for independent analysis. 
Public datasets have some associated disadvantages: they are either too generic for perception tasks or too task-specific~\cite{9156412}; they are extremely focused on the classes: car, pedestrian and cyclist~\cite{Teichman}; they are ``public'' only for research purposes, and cannot be used in the industrialization process of perception algorithms; most do not present data collected in different weather conditions; They are small, which prevents them from obtaining the minimum quantity of training data required to train neural networks to be competent in perception tasks. Virtual KITTI, SYNTHIA, Synscapes, Sintel and CARLA are examples of synthetic data datasets. On the other hand, some datasets that use real data are BDD, KITTI, ScanNet, NuScenes and ACDC~\cite{nowruzi2019real}\cite{synthetic}\cite{9711067}. The Middlebury flow dataset contains both real and synthetic scenes~\cite{4408903}. 

Most datasets focus on 2D annotations for RGB camera images. CamVid, Cityscapes, D2-City, BDD100k, Apolloscape and ACDC have released datasets with segmentation masks (image processing method in which a small 'image piece' is defined and used to modify a larger image). Vistas, D2-City and BDD100k complement their datasets with images taken during different weather conditions and illumination. ACDC~\cite{9711067}, the Adverse Conditions Dataset with Correspondences, consists of a medium-sized set of 4006 images which are equally distributed between four common adverse conditions: fog, nighttime, rain, and snow. ACDC supports both standard semantic segmentation and uncertainty-aware semantic segmentation. Multimodal datasets (consisting of images, sensor data, and GPS data) are expensive and difficult to collect due to the difficulty in synchronizing and calibrating sensors. KITTI was the pioneer in multimodal datasets combining dense point clouds provided by the LiDAR sensor with front-facing stereo images and GPS/IMU data. It was a great help in advancing 3D object detection. Thereafter, multimodal dataset KAIST uses colour and thermal cameras and a beam splitter to capture the aligned multispectral (RGB colour + Thermal) images. Thus, the dataset is able to provide data during the night capturing various regular traffic scenes, but the annotations are in 2D~\cite{9156412}. Other notable datasets are: LiDAR-Video Driving benchmark dataset which is among the first attempts to utilize point clouds to help driving policy learning and provide driving behaviour labels~\cite{8578713}; two 3D outdoor datasets presented by Hojung Jung et al. for semantic place categorization labels: forest, coast, residential area, urban area and indoor/outdoor parking lot~\cite{3Doutdoor}; and Málaga Urban Dataset gathered entirely in urban scenarios providing raw data without semantic labels~\cite{Blanco_Claraco2013Oct}.

Several studies that discuss the pros and cons for synthetic and real data can be found in the literature. In 2018, Tremblay et al.~\cite{tremblay2018training} presented a system for training deep neural networks for object detection using synthetic images. In order to force the network into learning only the essential features of the task, they use domain randomization for car detection. The idea was to effectively abandon photorealism in the creation of the synthetic dataset. This study also proved that, using real images, the accuracy of the models improves. In contrast, ~\cite{alhaija2017augmented} propose an alternative paradigm combining real and synthetic data for learning semantic instance segmentation and object detection models. The authors conclude that cluttering the images with too many objects reduces the model performance, and models trained on augmented imagery generalize better than those trained on synthetic data. Grand Theft Auto (GTA) game is used in the article by~\cite{tremblay2018training} to propose a fast synthetic data generation approach. The authors demonstrate that a state-of-the-art architecture, which is trained only using synthetic annotations, performs better than the identical architecture trained on human annotated real-world data.
In the context of LiDAR sensors, Wu et al.~\cite{wu2017squeezeseg} employ GTA combined with the KITTI dataset in order to create a synthetic LiDAR dataset to train a deep model and synthesize large amounts of realistic training data. Thereafter,~\cite{fang_2020} propose a novel LiDAR simulator that augments real point cloud with synthetic obstacles. First, using a LiDAR sensor and a camera, a real background environment dataset is created. This data is then augmented with synthetic 3D objects. The authors conclude that mixing real and simulated data can achieve over 95\% accuracy. 

Another relevant topic is the structure of the dataset.
The structure of data is an intriguing topic for data organization and representation. Choosing whether to employ structured, semi-structured, or unstructured data can have a significant impact on a project’s success. Fischer et al.~\cite{ijgi9110626} present a research data management system that includes a structured data storage for spatio-temporal experimental data in their study. Because data must be free, accessible, interoperable, and reusable, the usage of structured data is recommended. 

\subsection{Driveability Factors}
\label{sec:driveability}
Understanding what factors influence a scene's drivability is critical for analysing the variables collected during a simulation. Environmental factors like weather, traffic flow, road quality, and road obstructions, among others, are recognized to have a significant impact on the driving environment. However, these explicit characteristics, i.e. that can be immediately observed from the environment, are insufficient for evaluating the driving environment. The implicit information that must be inferred from observation must be taken into account. According to U.S. Department of Transportation reports,~\cite{guo_paper} considered important factors for driving based on studies of driveability in other fields of transportation systems research, U.S. Department of Transportation reports, and industry standards for estimating road risk. The explicit factors considered in this study were:
\begin{itemize}
    \renewcommand{\labelitemi}{$\bullet$}
    \item \textbf{Weather:} bad weather conditions, such as fog, rain, and wind, can limit a driver's road visibility. As a result, detecting objects and barriers can be challenging. \ac{DNN} models, for example, have a history of misbehaving in bad weather~\cite{weather_paper}.
    \item \textbf{Illumination:} perception is challenged by fluctuations in brightness induced by the time of day, the landscape, and directed light sources. The nighttime environment presents extra difficulties due to low illumination, changing contrast, and less colour information. As a result, studies on nighttime data are underrepresented~\cite{night_dataset}.
    At night, the vehicle's headlights and taillights help drivers recognize road objects. However, other illuminant sources such as traffic lights, street lamps, and road reflector plates on ground can cause many difficulties for detecting actual vehicles~\cite{vehicle_lights}. However, because pedestrians and other barriers lack their own light, it becomes difficult to identify them. Training in night situations reduces the accuracy of pedestrian detectors, according to a study conducted by~\cite{night_dataset}. 
    \item \textbf{Road Geometry:} it is significantly easier to drive on free ways or straight roads. Because of the high number of accidents that occur in these places, road designs such as intersections and roundabouts are extensively examined. Shirazi and Morris's~\cite{shirazi_morris} research examines recent studies on vehicle, driver, and pedestrian behaviour at intersections, as well as their levels of safety.
    \item \textbf{Road Condition:} the road's condition may be affected by its uneven surface, road damage, potholes, or construction. Since these examples are not very common, there is a lack of labelled data in this field. Construction on the road can alter the driving environment by adding traffic signs, changing the geometry of the road, and workers on the road (pedestrians). In this regard,~\cite{workzones} provide a set of computer vision methods that recognize the limits of a road work zone as well as transitory changes in driving surroundings. This restriction is important in order to determine the available and safe area for driving while avoiding potential risks.
    \item \textbf{Lane Marking:} the detection of lanes or roadways is made feasible by lane markings. There are several roadways that have no lines or have irregular lines. This makes detecting and delineating highways a difficult process. Some research, such as that conducted by~\cite{lane_mark1} and~\cite{lane_mark2}, show how to discover unmarked roads. These experiments aim to combine numerous inputs from cameras, infrared sensors, and LiDAR sensors.
    \item \textbf{Traffic Condition:} there is a distinct contrast between driving in an urban area and driving outside of an urban area. Some of the elements that distinguish the two driving environments are speed limits, traffic flows, the number of lanes, and traffic rules.
    \item \textbf{Static and Dynamic Objects:} one of the most researched topics in autonomous driving is object perception and detection. Existing approaches still have high error rates when it comes to finding things that are occluded by others, are small, or are difficult to recognize. These things can make driving difficult and inhibit effective planning. Pinggera et al.~\cite{lost_and_found} provide a dataset with small objects and a stereo vision algorithm for reliably detecting such impediments as lost cargo from a moving vehicle. Also,~\cite{Zhang_2021}, as well as~\cite{vanet}, built a framework for decision-making in driving situations with hidden agents in their research.
\end{itemize}

The actions and intentions of road users constitute implicit elements. Road users communicate with the autonomous vehicle and other road users. There are three realms to consider: the vehicle's interior, its surrounding environment, and the interiors of other vehicles~\cite{implicit_factors}. \cite{Sadigh-RSS-16} conducted a relevant study on this topic that reveals how autonomous vehicles' actions affect the responses of other vehicles.
Some implied factors are: \textbf{vehicle behaviour} when overtaking, lane changes, speed-driving, non-compliance with traffic laws, and other harmful vehicle behaviours are evaluated; the \textbf{behaviour of pedestrians}, which represents the most vulnerable road users. Many of the accidents involving autonomous cars also involve pedestrians. Rasouli and Tsotsos' study outlines many methodologies for studying pedestrian behaviour, as well as two approaches for predicting pedestrian intent. The first is to approach it as a dynamic object tracking issue, calculating the future trajectory of pedestrians, while the second is to approach it as a classification problem, categorizing pedestrian behaviour as “crossing” or “not crossing”. Methods that rely solely on pedestrian position are prone to errors. Other criteria, such as age, gender, and speed, must also be considered in order to forecast pedestrian intent and thus minimize collisions and other incidents~\cite{pedestrians_behaviour}; \textbf{driver behaviour}, since their intervention is still required in automatic driving when it fails or the car is unable to make trustworthy decisions. When investigating the causes of traffic accidents, elements such as skill, intention, driving style, distraction, and others are considered~\cite{driver_inattention}. Various ways for identifying factors such as driver tiredness and distraction while driving are shown in studies such as those conducted by~\cite{face_expression2} and~\cite{face_expression1}. Visual elements such as face expression and eye movement are used in traditional techniques. Autonomous vehicles currently have technologies in place to recognize when a driver is tired or preoccupied. Bosch's driver drowsiness detection, for example, monitors steering movements and urges drivers to have a rest when necessary~\cite{driver_drowsiness_detection}. 

\subsection{Overtaking Factors}
Several computer models are nowadays able to outperform humans in detecting and identifying objects, both in images and videos. However, autonomous vehicles, in addition to recognizing their external surroundings, have to make decisions, which has implications regarding safety, performance, ethics, and accountability. Mistakes during decision-making can result in severe accidents. Every year, road accidents result in about 20-50 million injuries and 1.25 million deaths. Many of these accidents are due to driver misinterpretation and untimely decisions, wrong speed choice, not being able to see through an obstacle, sudden breaks, ignoring road conditions, adverse weather conditions etc~\cite{PERUMAL2021104406}.

The overtaking manoeuvre is a way for faster drivers to continue driving at the desired speed without lagging behind slower vehicles. It brings comfort to the driver and enhances their experience on the road. However, that manoeuvre is one of the most critical actions in driving. The process involves lane-changing, acceleration and deceleration actions, calculating the distance of the oncoming vehicle and speed of overtaking and overtaken vehicles. 
To reduce the impact of these manoeuvres and increase driver safety, the vehicles would have built-in intelligent algorithms that consider all important aspects during decision-making. These aspects can be: calculating the proximity of other vehicles to the ego vehicle, determining whether a lane change manoeuvre can be made, and designing optimal and safe paths for the manoeuvre.

Most works and datasets mentioned are built mostly with respect to object detection and environment perception. However, overtaking scenarios must consider factors other than those mentioned. The research of \cite{DIXIT201876} considers the variations in number of overtaken vehicles, the duration of the overtake, the relative velocity between concerned vehicles and the distance between concerned vehicles. Features are classified as permanent (road and lane limits), slowly changing (speed limits, road works, traffic density, etc.), and fast changing (surrounding vehicle velocity, position, heading, etc.). According to the study, two crucial parts of high-speed overtaking trajectory planning are the integration of vehicle dynamics and environmental restrictions, as well as precise information of the surrounding environment and impediments. The research also proves that autonomous cars must have a precise understanding of the surrounding environment, which is not representative of real-world driving. Another good example is the study conducted by~\cite{5475195}. Some different features considered were lane markings, velocity and yaw rate, position and heading of the vehicle, longitudinal acceleration, distance  between vehicles, time to predicted collision and deceleration to safety time. They present a system that can perceive the vehicle’s environment, assess the traffic situation, and give recommendations about lane-change manoeuvres to the driver.

~\cite{901899} investigated on the minimum longitudinal distances required for lane changes or merging. For a vehicle overtaking another slower vehicle in front of it, \cite{1284727} described the equations of motion employed and the ideal values of the variables. \cite{5971279} proposed a fuzzy logic decision control system that achieves two consecutive lane changes. \cite{6629642} built a Bayesian belief network to calculate the probability of crashes in a driving environment with one car in front and one behind.

Certainly the greatest difficulty in overtaking is, during a fast flow of traffic, estimating the time available to perform the manoeuvre. Driver's decisions are unpredictable, especially when there is a speed difference between fast- and slow-moving vehicles. 
In their study, \cite{ASAITHAMBI2017252} considered the flying (with no other vehicles nearby, the vehicle overtakes a slow moving vehicle without having to slow down) and accelerative (the vehicle approaching a slow moving vehicle reduces its speed until it has enough space in the other lane to overtake) overtaking scenarios. The features analysed were the acceleration characteristics, speed of the overtaking vehicles, overtaking time, overtaking distances, safe opposing gap required for overtaking, flow rates, overtaking frequencies, types of overtaking strategy, and types of overtaking and overtaken vehicles. The authors proved that the majority of vehicles are travelling with their current speed without reducing the speed during overtaking (flying performed by 62\% of drivers and accelerative by 38\%). 

Another work that goes towards decision-making in overtaking scenarios is the work conducted by~\cite{SBS_Yang}. The authors designed a \ac{DNN} model to make overtaking decisions for stationary  vehicles and analysed the significant decision factors for these cases. They demonstrated that the factors extracted during the process of evaluating the traffic scene helped to improve the model's learning performance. The factors considered were divided into 3 categories: A – preceding vehicle, B – surrounding vehicles and C – ego vehicle. The factors included in category A were lateral distance to the right/left boundary of the road; duration of time it has been detected as stationary; velocity; acceleration; yaw angle; yaw rate; lane occupancy rate and object width/length. Category B included position, velocity, and acceleration to the nearest vehicle and number of vehicles, free space rate, spatial gaps and time/space mean speed for the remaining vehicles. Category C included relative speed, relative distance, time-to-collision/time-headway, and waiting time duration. All relevant factors were chosen based on the general characteristics of human drivers. 
Although all of these factors appear to be relevant in decision-making, in order for the model to be more efficient, the most crucial factors have to be chosen. The \ac{SBS} approach was used to determine the relevance of each factor. This method evaluates the learning performance of a model that takes all candidate factors as input to estimate the importance of each factor. It assesses the decrease in learning performance by eliminating a candidate. The significance of the factor that has been eliminated is determined by the degree of learning performance decrease. The method iterates the process for all factors, removing the least important one each time. The deleted factor is regarded unimportant or redundant if the drop is almost zero or minor. In short, the SBS technique determines the dominant set by analysing the relevance of the features.
As a result, the lateral position of the previous vehicle was found to be the most crucial component, followed by the vehicle's waiting time to begin the overtaking manoeuvre. This indicates that the driver's decision can change over time, even though the situation remains the same. In the end, factors like time/space mean speed and lane occupancy rate were found to have no discernible effect on performance. Table~\ref{fig:sbs_yang} shows the outcome of the SBS method applied to the factors in this research.

\begin{table}[H]
    \centering
    \caption{Outcome of the SBS method applied to the factors in the research by~\cite{SBS_Yang}.}
    \label{fig:sbs_yang}
    {\small
    \begin{tabular}{|c|c|c|c|} \hline
    Rank & Category & Feature Description & Accuracy (\%) \\ \hline \hline
    1    &  A       & Lateral Position    &  65.1 \\ \hline
    2    &  C       & Waiting Time        &  79.8 \\ \hline
    3    &  B       & Time Mean Speed     &  83.0 \\ \hline
    4    &  B       & Number of Vehicles  &  87.0 \\ \hline
    5    &  C       & Distance to Preceding Vehicle &  88.4 \\ \hline
    6    &  A       & Moving Confidence    &  89.1\\ \hline
    7    &  A       & Current Speed    &  89.4 \\ \hline
    8    &  B       & Speed of the Closest Vehicle    &  89.5 \\ \hline
    9    &  B       & Lane Occupancy Rate   &  89.5 \\ \hline
   10    &  B       & Space Mean Speed   &  89.5 \\ \hline
    \end{tabular}
    } 
\end{table}

\section{A dataset oriented to overtaking manoeuvering}
Synthetic data is artificially generated information that can replace real data when it lacks quality, volume, or variety. Real data is also sometimes insufficient when it does not meet the needs of what is intended, or when its creation may cause danger or damage. Synthetic data is widely used in the field of artificial intelligence to, for example, train models when real data is lacking, to fill gaps in training data, to predict the future (old data may lose its value), to generate marketing images, etc. Generating synthetic datasets that are statistically significant and relevantly reflect real data can be challenging, since it is necessary to guarantee that the generated data is similar enough to the real data.
Synthetic data is an added value for training neural networks, as they are more accurate when trained with a wider variety of data. However, gathering and labelling such massive datasets with thousands or even millions of objects is too expensive. In this regard, synthetic data can save money since simulators can reproduce scenarios that are challenging to reproduce in the real world.
Apart from bringing the opportunity of collecting large amounts of data that in the real world may not be possible due to time limitation (there may not be time to collect the required amount of data), synthetic data also adds the advantage of being able to collect dangerous data that in the real world may cause harm and/or put a living being at risk.
A synthetic dataset should be sufficiently diverse, but it is necessary to control the randomness of data generation in order to make it realistic. To generate synthetic data, models produce synthetic data based on the probability that certain data points will appear in the real dataset. Neural techniques like Variational Autoencoders and Generative Adversarial Networks are commonly used to generate synthetic data. In autonomous driving, synthetic data is often generated with the help of car simulators. The synthetic data is simulated data that generates photorealistic simulations that follow the laws of physics. Simulated data includes all necessary annotations and dimensions, producing realistic 3D data.

\begin{figure}[H]
  \centering
  \includegraphics[width=0.6\linewidth]{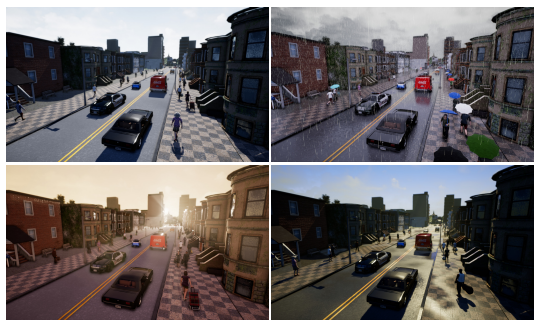}
  \caption{Third-person view in four weather conditions~\cite{Dosovitskiy17}.}
  \label{fig:carlaweather}
\end{figure}

The simulator chosen in this work was CARLA, an open-source tool for autonomous driving research that supports the development, training, and validation of autonomous urban driving systems. 
Following a thorough evaluation of numerous driving simulators, CARLA emerged as the most notable for being highly complex, having a variety of useful functionalities for this work, including the incorporation of weather conditions, and providing a thorough and understandable documentation.
CARLA has a flexible API that allows users to control all aspects related to the simulation, including traffic generation, pedestrian behaviors, weather, sensors, and much more~\cite{Dosovitskiy17}.
The platform includes a large set of objects as well as static and dynamic actors that may be used to customize maps and scenarios. A wide range of sensors can be added to the car to collect information about all environment; weather can be customized in detail (Figure~\ref{fig:carlaweather}); simulations can be recorded and played and a blueprint library with a wide set of actors can be used.

The simulator makes it feasible to create and manipulate every element of the road (something impossible in the real world), including the  road layout, the flow of traffic, the types of vehicles present, assign speed values and perform manoeuvres such as lane changing. It is also possible to create risky circumstances using the simulator, such as collisions or running over. In order to collect as much information as possible and realistic, a planning and structuring phase of the scenarios and variables to be collected precedes the implementation of the simulations. 

\subsection{Limitations of the Simulator}\label{chap:limitations}
Despite its usefulness, ability to simulate complex scenarios and popularity, the CARLA simulator exhibits some limitations. When attempting to complete two lane changes in a row (in considerably close frames), the first limitation was encountered. 
The responsibility for the lane change process lies within the Traffic Manager. When the method ``force\_lane\_change(self, actor, direction)'' is called, a lane change is done to the vehicle described in the actor to the right if direction is True and to the left if it is False~\cite{traffic_manager}. This method has an associated timeout that prevents the execution of a new job before the old one has completed. To tackle this issue, two TrafficManagers were employed. Upon the first lane change of the ego vehicle, the first TrafficManager is turned off, thus ending its tasks. To make the second lane change and return to its original lane, a second TrafficManager is turned on and paired with the vehicle. Often an overtake can occur at high speeds and consequently take a few seconds between the first and the third stage (explained in Subsection~\ref{chap:collected_data}). Without this solution, these circumstances, and data could not have been represented or collected.

Another crucial limitation of the system is the impossibility to force a lane change with the Traffic Manager when the vehicle is at a speed higher than 40 km/h. We propose different solutions to this problem. One approach is to reduce the speed to below 40 km/h the moment the car starts the overtaking manoeuvre, and then return to the previous speed. This strategy was quickly shown to be impossible, since it manipulated/changed the simulation far too much. The values recorded were far from a realistic scenario. Because the vehicle was travelling at such a slower speed than the other vehicles, there was no risk of crashes or infractions of different traffic laws. It was realized that the speed difference was important for the preservation of the simulation environment and could not be changed. With this, another approach was taken where a conversion of the randomly generated speeds from a simple rule of 3 was performed. The maximum speed considered in the simulations is 120 km/h, so this value would correspond to 40 km/h in the simulator (maximum speed to be able to overtake). Once the speeds are converted, the timestamps are also converted and the differences between the vehicle speeds are preserved. 

Finally, for the weather conditions, the simulator provides a visual representation of various weather conditions: cloudy, rainy, day/night, sunset, etc~\cite{carla_weather}. However, the physics of the car during the simulation does not change depending on the level of precipitation, wind or fog. 
The same happens with the time of the day. The simulation does not differentiate between day and night. There is no reduction in road visibility and no activation of vehicle lights. The system has been updated to reflect these two modifications.
The influence of weather conditions was considered an asset to the work. Therefore, since it is not implemented in the system, it was necessary to apply forces to the car depending on the level of rain and wind. Depending on the wind direction, a force is applied to the car to move it to the right or left. Forces are applied at longer time intervals for a lower average wind and rain values, and vice versa. When a high level of cloudiness (above 60\%) is registered, or when the sun is at the driver's eye level (horizon line), his field of vision is reduced.  

\subsection{Scenarios}\label{chap:scenarios}
The first step in generating the data corresponding to overtaking situations is to create the scenarios that, by common knowledge, are considered possible to happen. 
In order to obtain positive, negative, and neutral cases for model training, successful, unsuccessful, and non-overtaking scenarios were considered. All environments are created randomly, but ensuring that all cases are represented. First, a random set of cars are placed in a portion of the world. The location, type, colour, and speed of the vehicle as well as weather conditions are randomly generated for each simulation.

The number of cars in each simulation varies from 2 to 6. A minimum of 2 guarantees there is always an ego vehicle and the vehicle it intends to overtake. A maximum of 6 was chosen considering the available space on the chosen portion of the street.
Each vehicle is assigned initial values of location x, y and z and speed ranging from 50 to 120 kilometres per hour. The values were chosen because they represent the lowest and highest speeds permitted on Portugal's highways in article 27\ts{th} of the Portuguese Highway Code \textit{(Código da Estrada)}~\cite{velocity_ce}. During the simulation, these values may fluctuate and may surpass 120 km/h or fall below 50 km/h. Various vehicle types were considered in the simulations, including small, medium and large light passenger vehicles, vans, trucks, bicycles, and motorcycles. Bicycles or motorcycles can not be the ego vehicle, since our focus is on autonomous light passenger vehicles. Another distinguishing characteristic of the cars is their colour. Yellow, blue, red, green, orange, grey, white, and brown are all options.

\begin{figure}[H]
\centering
\begin{subfigure}[t]{.24\textwidth}
    \centering
    \captionsetup{justification=centering}
    \includegraphics[width=.95\linewidth]{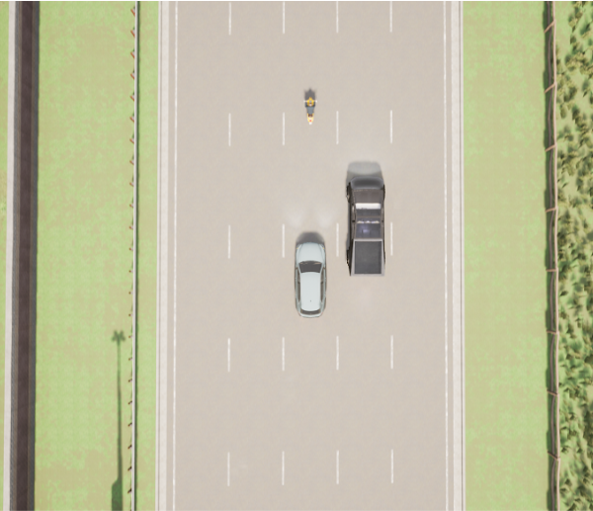}  
    \caption{Clear Noon \\ Weather}
    \label{fig:day_sim}
\end{subfigure}
\begin{subfigure}[t]{.24\textwidth}
    \centering
    \captionsetup{justification=centering}
    \includegraphics[width=.95\linewidth]{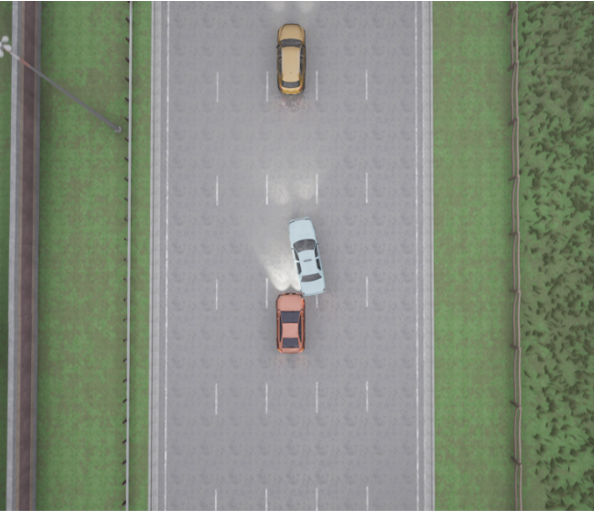}  
    \caption{Hard Rain Noon \\ Weather}
    \label{fig:rain_sim}
\end{subfigure}
\begin{subfigure}[t]{.24\textwidth}
    \centering
    \captionsetup{justification=centering}
    \includegraphics[width=.95\linewidth]{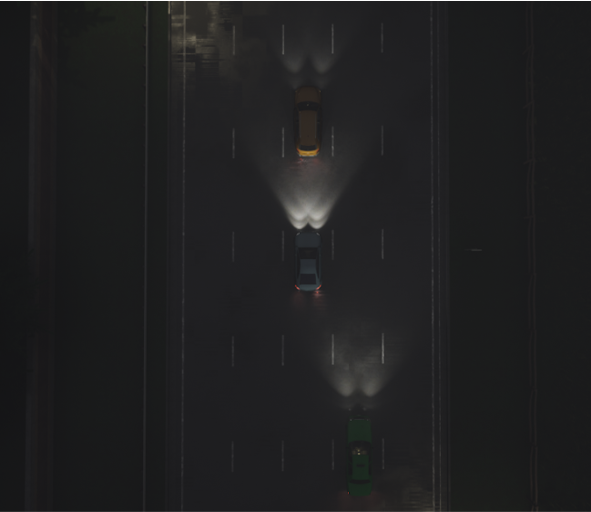}  
    \caption{Clear Night \\ Weather}
    \label{fig:nigth_sim}
\end{subfigure}
\caption{Overtaking scenarios with multiple weather conditions.}
\label{fig:scenarios}
\end{figure}

Finally, each simulation has its own set of meteorological conditions. Cloudiness, precipitation, wind intensity, and sun height angle are among the characteristics that can be changed in the simulator. For convenience, CARLA also provides a list of predefined weather presets that can be directly applied to the world. As a result, this work's simulations can select among the following weather conditions: SoftRainNight, MidRainNight, or HardRainNight, which depict a nighttime scenario with low, mid, and high percentages of precipitation and fog, respectively, as well as low wind intensity. HardRainNoon indicates a daytime scenario with strong precipitation, and low wind and fog intensity; ClearNoon represents a daytime scenario with no precipitation, and low wind and fog intensity; ClearSunset represents a daytime scenario with a horizon line (sunlight at the driver's eye line) without precipitation and low wind and fog intensity; ClearNight represents a scenario with the same aspects as ClearNoon but at night; and CloudyNight represents a nighttime scenario with high intensity of fog and low precipitation and wind intensity~\cite{weatherparameters}. An example of simulations performed in the ClearNoon, HardRainNoon and ClearNight scenarios can be seen in Figures~\ref{fig:day_sim},~\ref{fig:rain_sim} and~\ref{fig:nigth_sim} respectively.

\subsection{Collected Data} \label{chap:collected_data}
At each frame of the simulation, relevant information regarding each object present is collected. The data is entered into a data table, and Table~\ref{tab:data_collected} displays the description of the stored variables. 

\input{tables/variables}

The variables collected were chosen based on the literature and on the Portuguese road traffic, the document that regulates the traffic of people and vehicles since 1901 in Portugal~\cite{acp}. It has been proven, as mentioned in Section~\ref{sec:driveability}, that factors such as weather, illumination, and road geometry, among others, influence driving, making these variables also essential in overtaking scenarios. 
The variables collected in each simulation can have qualitative/categorical and quantitative/numeric values. 

Each moment of overtaking is represented by the variables S, F, and TS. S stands for the simulation's unique identifier, F for each frame's unique identifier, and TS indicates the seconds passed from the beginning of the simulation to the moment it is recorded. Information such as each vehicle's unique identity (IDego for the ego vehicle and idv for the others) and dimension (Dim) are saved to identify the cars present in the simulation. To monitor all the activity of each vehicle, the location (L), recorded speed (V), wheel direction (D) and acceleration (A) are stored for each vehicle in each frame. The vehicle's location is given by a point x,y together with the unique identification of the lane the vehicle is in. To account for the traffic condition factor, the maximum speed value (MV) for the lane in which the ego vehicle is located is recorded. It's possible to tell if it is exceeding the maximum speed by comparing its current speed (V) to the
maximum speed (MV). Considering other factors such as road geometry and lane marking, the types of lines on the right (RT) and left (LT) of the ego vehicle are recorded, as well as the width of the lanes that surround it (LW for the lane it is in, LWR for the lane on its right and LWL for the lane on its left). These variables are crucial to check, for example, whether the vehicle crosses a solid line (which is prohibited by Portuguese Highway Code in article 146\ts{th}~\cite{line_ce}). 
In turn, by comparing the variables car size (D) and road width (LWR and LWL), it is possible to determine if the lane in which the automobile is heading has sufficient width. 
Since collisions between vehicles constitute a danger scene, it is necessary to store this information (variable C). Weather and lighting are also taken into account. These restrict the driver's view and make it harder to interpret the surroundings. As a result, rain (Prec), fog (Fog), and wind (Wind) percentages are measured. The time of day is stored in the variable DN which can register whether it is day or night. HL records whether the horizon light is dazzling the driver, a factor that is also considered relevant in limiting driver visibility.
Finally, the variable OV simply stores the moment when the vehicle starts the lane change manoeuvre. This variable can be used to determine how many overtaking phases the vehicle has completed. Considering v as the car that the ego vehicle intends to overtake, three stages are considered in a successful overtake: the first is before the ego vehicle executes the lane change (ego vehicle is behind v in the same lane), the second is when it has already changed lanes and tries to reach a position ahead of v; and the third is when the ego returns to the initial lane, leaving v behind. 
So, if there are two ones in OV in a simulation, the vehicle has moved from the first to the second stage and recorded a 1, and then went to the third stage and recorded another 1. If it just gets one value 1 in OV, it has only progressed from the first to the second stage and was unable to go back to its initial lane. There is no lane change if OV does not have any value 1. The category in these two last circumstances is considered a non-overtake attempt.

\subsection{Planning and Structuring}\label{chap:planning}
Most of the data held by businesses is unstructured. However, it is well known that this data requires structuring in order to be used in decision-making processes. For a successful comprehension and utilization of data, a planning phase is essential. Many artificial intelligence researchers agree that many tasks, including reasoning, planning, and decision-making, rely on a combination of three mental mechanisms: neural, symbolic, and probabilistic~\cite{nye_2021}~\cite{tenenbaum_dehaene_2021}~\cite{FRISTON2021573}. The symbolic component is used to represent and reason about abstract knowledge. The probabilistic inference model aids in the establishment of causal relationships between objects, the reasoning of counterfactual or never-before-seen events, and the handling of uncertainty. Finally, pattern recognition is used by the neural component to link real-world sensory data to knowledge. 
Data is a representation of facts, a simple observation of the world.
Data can be categorized into qualitative when it attributes a quality (e.g. colour of a vehicle) and quantitative when it can be measured (e.g. distance between two vehicles). However, it is necessary to interpret and contextualize the data in order to turn it into information. Structuring data is a crucial step in knowledge representation. This is how data is organized to be interpreted by machines and humans.  There are several data structures used in programming like lists, stacks, trees, graphs, hash tables, among others. Data models, on the other hand, are tools that allow to demonstrate how the data structures that will later support the decision processes are built. They explain how the data is organized and what relationships are intended to be established between them.

\begin{figure}[H]
    \centering
    \includegraphics[width=.5\textwidth]{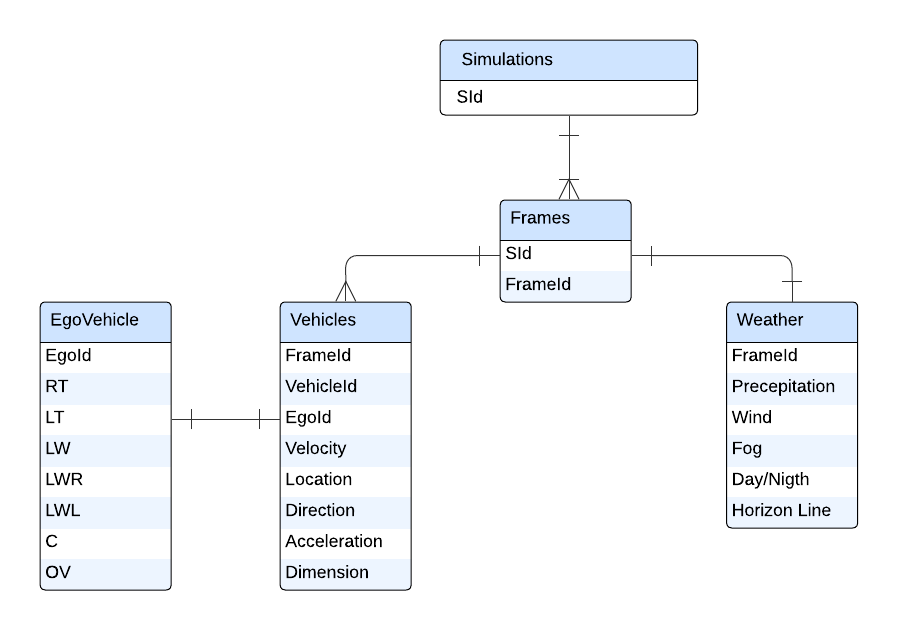}
    \caption{Entity-Relationship diagram developed.}
    \label{fig:erd}
\end{figure}

In this work, data from simulations are represented by an Entity Relationship diagram (Figure~\ref{fig:erd}). 
The ER diagram was chosen because it is a very well-known and understandable representation and because it provides a preview of how tables should link to one another and which properties should be emphasized or eliminated. Besides, it favours derived forms of knowledge representation such as first order logic or graph-based.

In the diagram, the different entities are represented: Simulations, Frames, EgoVehicle, Vehicles, and Weather. Each entity has its own attributes, for example, each vehicle (belonging to the Vehicles entity) has an associated speed.  Here the Ego Vehicle attribute is differentiated from the Vehicles attribute, which in turn stores information about all cars (including the ego vehicle), because the values of the variables RT, LT, LW, LWR, LWL, C and OV are only stored for the ego vehicle. The relations are also represented as well as their cardinality. An example is the relation between the entity Simulations and the entity Frames that is a one-to-many relation, that is, each simulation has one or more frames, but each frame only belongs to one simulation. 

\subsection{Feature Engineering}
Feature engineering is the process of transforming raw data into useful features that are later used by the algorithm.
This includes the process of creation, transformation, extraction, and selection. During this stage, features that add relevant information to the study and that can support the classification of the overtaking manoeuvre are created and included to the dataset. Because all data is acquired using the CARLA simulator, no information is gathered from any other source. Using the imputation technique, missing values are removed. In frames when no collisions are recorded, missing values are discovered. Since the values are numeric, the missing values are set to 0. Following the creation process, each feature's statistics are analysed by computing the minimum, average, and maximum observed value for each class, as well as the standard deviation. 

In the process of creating the features, features created in different important studies in the literature are taken into account. The features are divided into two categories: static (values that don't change over time) and dynamic (values that change in each frame). Let E be the ego vehicle and P the vehicle that E intends to overtake. The static features considered are: time of the day; presence of horizon line; type of E; type of P; waiting time (time that E waited to start the overtaking manoeuvre); overtaking time (total time the overtaking manoeuvre takes); number of vehicles present in the simulation.
The dynamic features considered are: current speed of E; current speed of P; speed difference between these two vehicles; distance between E and P; occupancy rate of the E's left lane (the lane E intends to head towards); weather, percentage of precipitation, wind and fog.
Tables~\ref{tab:static_res} and~\ref{tab:dynamic_res} illustrate the static and dynamic features created, respectively, as well as the measurement units of each. Letters S, M, L, V, T, B and MC represent the possible types of vehicles present in the simulation. S, M and L represent the small, medium and large light passenger vehicles, respectively; vans, trucks, bicycles, and motorbikes are represented by V, T, B and MC, respectively.

\begin{table}[H]
       \caption{Static Features Created.}
      \label{tab:static}
      \centering
      \resizebox{.4\textwidth}{!}{%
        \begin{tabular}{@{}|c|c|c|@{}}
        \toprule
        \rowcolor{blue} 
        Name & Static Features & Measurement \\ \midrule
        DN & Time of the Day & Day or Night  \\ \midrule
        HL & Horizon Line & Yes or No  \\ \midrule
        TE & Type of E & S, M, L, V or T \\ \midrule
        TP & Type of P & S, M, L, V, T, B or MC \\ \midrule
        WT & Waiting Time & seconds \\ \midrule
        OT & Overtaking Time & seconds \\ \midrule
        NV & Number of Vehicles & int number \\ \bottomrule
        
\end{tabular}%
        }
\end{table}

\begin{table}[H]

      \centering
        \caption{Dynamic Features Created.}
        \label{tab:dynamic}
        \resizebox{.4\textwidth}{!}{%
        \begin{tabular}{@{}|c|c|c|@{}}
        \toprule
        \rowcolor{blue} 
        Name &  Dynamic Features & Measurement \\ \midrule
        SE & Current Speed of E & kilometers per hour \\ \midrule
        SP & Current Speed of P & kilometers per hour \\ \midrule
        DSEP & Speed Difference Between E and P & kilometers per hour \\ \midrule
        D & Distance Between E and P & metters \\ \midrule
        OLR & Occupancy Rate of the E's Left Lane & percentage \\ \midrule
        PREC & Precipitation & percentage \\ \midrule
        WIND & Wind & percentage \\ \midrule
        FOG & Fog & percentage \\ \bottomrule
\end{tabular}%
        }
\end{table}

The features are created based on generally known overtaking scenarios along with information gleaned from study in the field. The selected factors are those considered to be relevant in the overtaking decision at the moment the driver initiates the manoeuvre. Time of day and horizon line are chosen as features because they contribute to a diminished perception of the road. As referenced in Section~\ref{sec:driveability}, researches show that night scenes make it difficult to perceive road objects and reduce the driver's understanding of the surrounding environment. The horizon line was an added feature because it represents a decrease in the field of view by being a light directed at the driver's eyes. In turn, the type of vehicles that overtake as well as those that are overtaken are also taken into account. The physics of a heavy vehicle differ from those of a light vehicle. Because it is heavier, it slows down and may take longer to complete the driving manoeuvre. A light vehicle that wants to pass a heavy vehicle, on the other hand, will need more time to reach a greater x-axis point than the other vehicle. The waiting time value is also essential, since a driver may be enticed to overtake if he has been driving behind that automobile for an extended period of time. 
The vehicle may have had enough time to adjust its speed to that of the vehicle in front of it and no longer sees the need to overtake. The overtaking time value represents the time it takes to manoeuvre from the moment the vehicle changes to the left lane until it returns to its initial lane, overtaking the desired vehicle. This aspect might also impact a driver's decision to overtake or not, if he knows how long the overtaking will take ahead of time. With a big safety distance between automobiles, a successful overtake will undoubtedly take longer. The number of vehicles in the simulation has an impact on traffic flow, which can play a role in the driver's decision to overtake. The motorist will have more confidence to overtake in a less congested environment, and the dangers of a failed overtake will be minimized.
In terms of dynamic characteristics, both the speed of the overtaking vehicle and the speed of the one being overtaken is important. A car driving at a high speed will act differently than one driving at a low pace. The values of reaction, braking, and stopping distance are affected by the speed values. These distances are affected not only by the vehicle's speed, but also by the condition of the tyres, the efficiency of the brakes, and the road surface (wet, slippery, or sandy)
. The condition of the car is not taken into account in this work. These values, as well as overtaking time and waiting time, are also influenced by the speed of the car being overtaken. Speed values may not be sufficient alone, so the difference between the speed of the overtaking car and the speed of the overtaken car was also considered. A greater difference in speeds can be the cause for overtaking to happen. In theory, an automobile travelling at a very high speed behind a car travelling at a very low speed will need to overtake.
Because the two automobiles are moving in the same direction and with the same orientation, the difference in speeds needs to be calculated. 
The distance between the two vehicles must also be taken into account. An important practice in driving is to increase the safety distance between cars as a way to prevent accidents.
Another factor chosen was the occupancy rate of the lane to the left of the vehicle (the one to which the ego vehicle wants to head to overtake the other car). The traffic flow must be considered once more, but this time in a specified lane. Finally, weather conditions are taken into account because, as is well known, they have an impact on the state of the pavement and the driver's visibility. In different weather circumstances, the identical overtaking scenario produces vastly diverse results. As previously stated, a scenario with heavy rain or fog can increase reaction time, braking distance, and stopping distance, resulting in crashes. Wind, in turn, might cause the vehicle's trajectory to be disrupted.

\subsection{Classification}\label{chap:classification}
\begin{figure}[H]
\centering
\begin{subfigure}[t]{.24\textwidth}
    \centering
    \captionsetup{justification=centering}
    \includegraphics[width=0.95\linewidth]{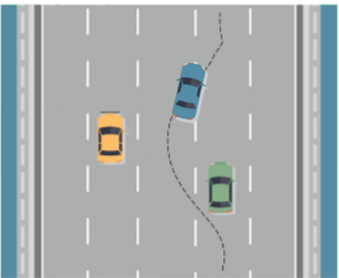}  
    \caption{Successful \\ overtaking}
    \label{fig:success}
\end{subfigure}
\begin{subfigure}[t]{.24\textwidth}
    \centering
    \captionsetup{justification=centering}
    \includegraphics[width=0.95\linewidth]{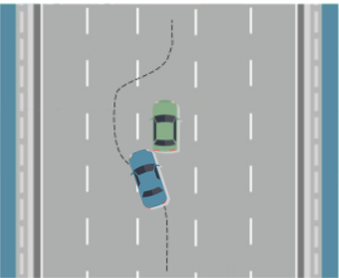}  
    \caption{Non-Successful overtaking (collisions)}
    \label{fig:bad_col}
\end{subfigure}
\begin{subfigure}[t]{.24\textwidth}
    \centering
    \captionsetup{justification=centering}
    \includegraphics[width=0.95\linewidth]{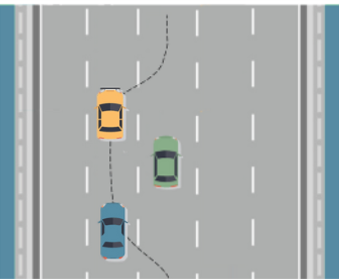}  
    \caption{Non-Successful overtaking (no collisions)}
    \label{fig:bad_no_col}
\end{subfigure}
\begin{subfigure}[t]{.24\textwidth}
    \centering
    \captionsetup{justification=centering}
    \includegraphics[width=0.95\linewidth]{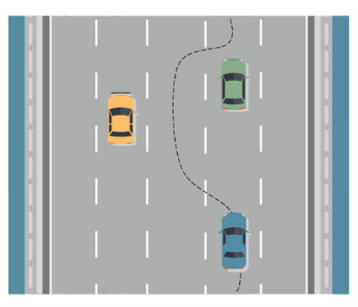}  
    \caption{No attempt to overtake}
    \label{fig:none}
\end{subfigure}
\caption{Classes chosen to represent the different overtaking scenarios. \\ Legend: The blue colour represents the ego vehicle; the green colour represents the vehicle that the ego vehicle wants to overtake; the yellow colour represents simple cars; the dashed black line represents the car's intended path.}
\label{fig:scenarios_classes}
\end{figure}

As mentioned in Subsection~\ref{chap:scenarios}, successful and unsuccessful overtaking scenarios are considered as well as non overtaking attempts. Let v be the vehicle to be overtaken by the ego vehicle. Scenarios in which the ego vehicle can achieve the following steps in order are classified as successful overtaking: visualize v (which is in front of it), ego vehicle change to the lane on its left; reach a location on the x-axis higher than the v; and return to its initial lane. Figure~\ref{fig:success} illustrates an example of a successful overtaking. In this case, the question is whether returning to one's initial lane is sufficient to be considered successful overtaking. A system that implements an overtaking decision method must additionally consider whether this presents a danger to other road users. According to article 38\ts{th} of the Portuguese Highway Code \textit{(Código da Estrada)}, a vehicle can only overtake another vehicle if the manoeuvre does not represent a danger to those passing on the road~\cite{back_lane_ce}. As a result, various traffic rules must be considered in order to assure the manoeuvre's legality as well as the driver's and other vehicles' safety. 
In light of this, successful overtaking was divided into legal and illegal successful overtakes.
In turn, overtaking considered unsuccessful are those where the ego vehicle begins the overtaking manoeuvre but does not finish. This can happen due to a collision or due to some event or obstacle that did not allow the manoeuvre to be completed, as illustrated in Figure~\ref{fig:bad_no_col}. In these cases, we might notice a decrease of speed after the car has started to overtake, which makes the car unable to get enough speed to overtake. Collisions, represented in Figure~\ref{fig:bad_col}, can have many causes, among them: poor visibility that prevents the driver from seeing the car ahead in time to brake safely; not keeping a necessary safety distance from the car ahead; adverse weather conditions that increase braking time; among others. Finally, the scenarios considered as neutral (no attempt to overtake) are those in which no overtaking manoeuvre is performed.

\begin{figure}[H]
\centering
    \includegraphics[scale=0.5]{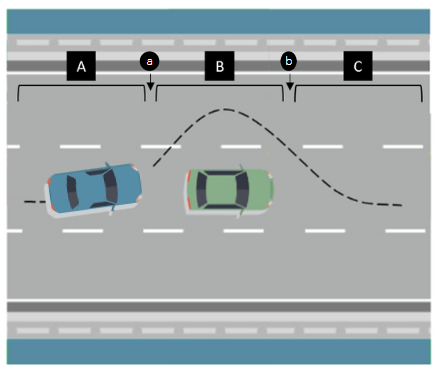}
    \caption{Representation of the 3 stages of a successful overtaking.}
    \label{fig:stages}
\end{figure}

Following that, a data analysis was performed to determine which elements had the greatest impact on the simulation's outcome. For each factor related to each class, the minimum, average, and maximum values, as well as the standard deviation, were gathered. These values are measured at the moment the driver initiates the overtaking manoeuvre. As indicated in Subsection~\ref{chap:collected_data}, a successful overtaking is one in which the vehicle reaches 3 stages. Figure~\ref{fig:stages} displays these three stages, as well as the transition points between them. The first step, represented by letter A, occurs when the ego vehicle remains behind the other it intends to overtake. The second stage begins when the ego vehicle starts the overtaking manoeuvre, which is symbolized by transition a in Figure~\ref{fig:stages}. The second stage continues with the vehicle in the lane it has moved into until it reaches an x-axis location larger than the other car. b represents the instant it executes the second lane change to return to its original lane. Thus moving on to stage 3, where the ego vehicle is in its initial lane, leaving the overtaken vehicle behind, represented by the letter C. As a result, all measures were taken at time a for successful overtakes, legal and illegal, as well as unsuccessful overtakes. For the non-overtake situations, the measured values in each frame of the simulation are averaged.

\subsection{Scenarios and Data Collected}
\begin{figure}[H]
\centering
    \includegraphics[width=0.3\linewidth]{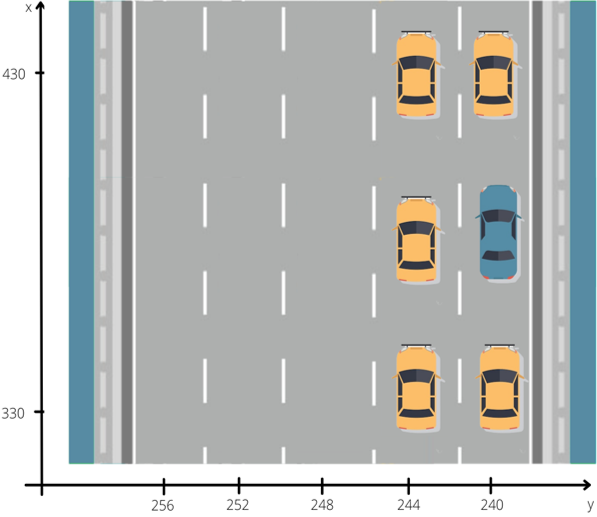}
    \caption{Street Portion Layout.}
    \label{fig:road_map}
\end{figure}

After planning and defining the scenarios to be represented, parameters were assigned at the beginning of each simulation. The number of vehicles as well as their type, colour, and speed were varied in the intervals already mentioned in Subsection~\ref{chap:scenarios}. The chosen portion of the world represents a one-way highway consisting of 5 traffic lanes. This portion has an x-axis between 320 and 450 and a y-axis between 238 and 258, as shown in Figure~\ref{fig:road_map}, where the ego vehicle is represented by colour blue and yellow vehicles are the surrounding ones. In the simulator environment, waypoints are represented by xyz axes. Because the z-points always have the value 0 in the measurements, they are ignored at the point where the cars have been added to the scenario (the cars are always on the ground). The value of z subsequently assumes significance, as the height of the cars can play a significant role in overtaking. The initial placement of the vehicles can be anywhere between 320 and 450 on the x-axis and on the y-axis each lane is represented by a value. That is, vehicles with a y-location between 238 and 242 are in the first lane, those with a y-location between 242 and 246 are in the second lane, those with a y-location between 246 and 250 are in the third lane, those with a y-location between 250 and 254 are in the fourth lane, and those with a y-location between 254 and 258 are in the fifth lane.
Since the ego vehicle always wants to overtake, it never starts on the fifth lane, since a free lane on its left is needed for it to perform the manoeuvre. \par

\begin{figure}[H]
\centering
    \includegraphics[scale=0.2]{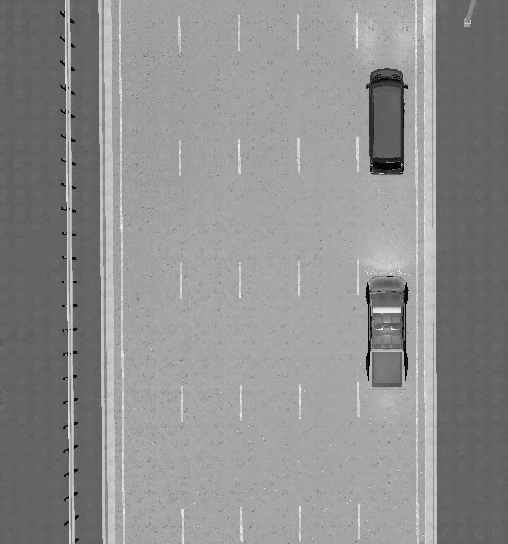}
    \caption{Screenshot of simulation 1 video at frame 70.}
    \label{fig:video}
\end{figure}

Once the parameters are assigned to each simulation, the output of each is recorded and organized in text format in a data table. As a crucial component for later analysis and interpretation of the data obtained in each simulation, an mp4 video is saved for each one. A screenshot of the video can be seen in Figure~\ref{fig:video}.

\subsection{Data Exploration}
The number of cases in the dataset corresponding to each class was calculated and illustrated with a histogram. All analysis in this section was performed using Python pandas and sklearn. 




Another representation for interpreting the data is the box plot, a tool that allows the visualization of outliers and the distribution of the data.
The diagram consists of a minimum and a maximum value and the first, second and third quartile. The median or second quartile is the mean value of the data; the first quartile is the mean value between the lowest number (not the minimum) and the median of the data; the third quartile is the mean value between the median and the highest value (not the maximum) of the data. Outliers are points of observation far away from other observations.
Since the variables were not all at the same scale, before computing next plots it was necessary to use the MinMaxScaler function provided by the sklearn.preprocessing package to transform the features by scaling each feature to a given range [0-1] by default. The dispersion of the data can be represented by the difference between the third and first quartile, i.e. the size of the box. The amplitude, in turn, is obtained from the difference between the maximum and minimum values. The dispersion is a more robust measure because it does not consider outliers~\cite{outliers2}.
A swarm-plot was also implemented where only the points are adjusted so they won’t overlap which helps with a better representation of the distribution of values~\cite{swarmplot}. 

\begin{figure}[H]
\centering
\sbox{\measurebox}{%
  \begin{minipage}[b]{0.5\textwidth}
  \subfloat
    []
    {\label{fig:pairplot}
    \includegraphics[width=0.9\textwidth]{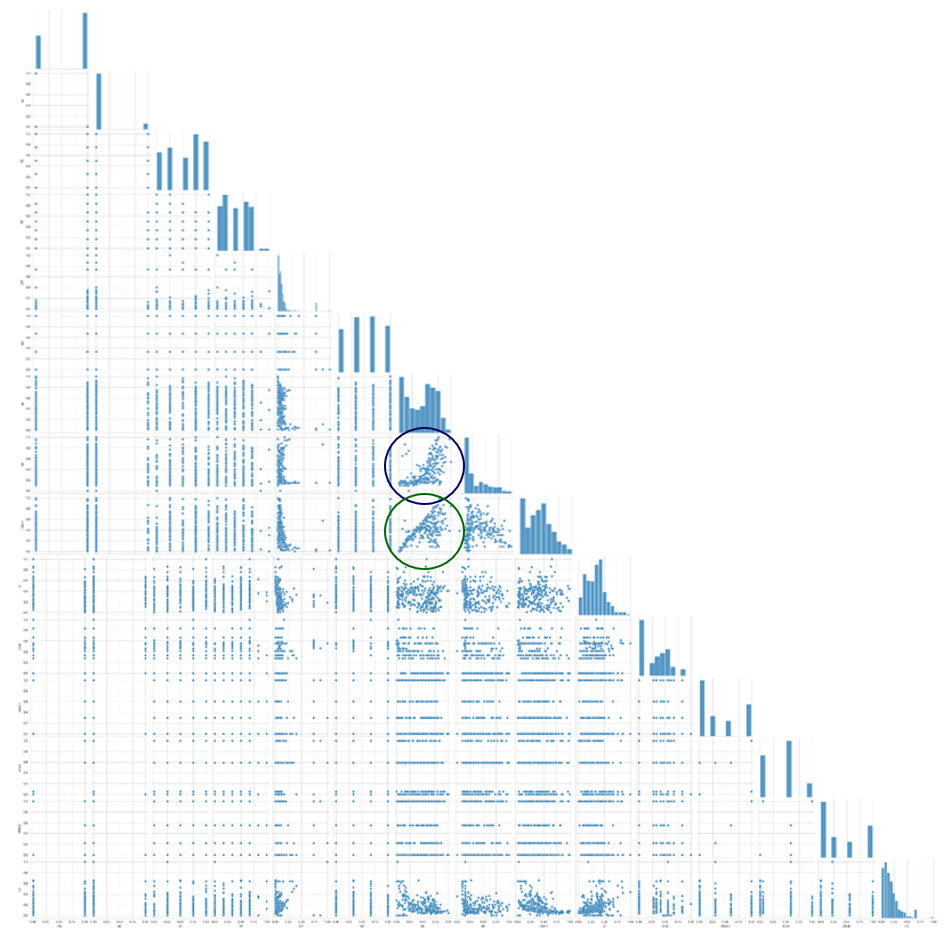}}
  \end{minipage}}
\usebox{\measurebox}\qquad
\begin{minipage}[b][\ht\measurebox][s]{.3\textwidth}
\centering
\subfloat
  []
  {\label{fig:SP_SE}\includegraphics[width=0.6\textwidth, cfbox=plot_blue 1pt 1pt]{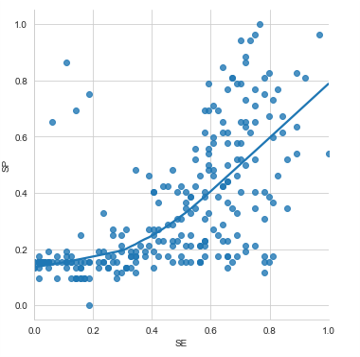}}

\vfill

\subfloat
  []
  {\label{fig:DESP_SE}\includegraphics[width=0.6\textwidth, cfbox=plot_green 1pt 1pt]{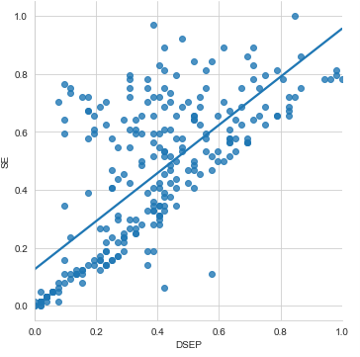}}
\end{minipage}
\caption{Features Relation. \\
    Legend: (a) corresponds to pair plot output for all features; 
    (b) corresponds to correlation points between SE and SP; 
    (c) corresponds to correlation points between DSEP and SE.}
\end{figure}

An interesting next step is to understand the relationship between two variables. For example, it would be desirable to know if the type of vehicle being overtaken has an impact on the ego vehicle's safety distance. A correlation coefficient is one technique to quantify this relationship. The degree to which two variables are linearly related is known as correlation. Correlation does not always imply causation, two variables can have a strong correlation due to a random exogenous occurrence. For example, a clothing store increases the number of sales of fur coats during the winter. As a result, there is a strong correlation between coat unit sales. It can be seen that there is a causal relationship in this example, as extreme winters enhance coat sales. Coat sales, on the other hand, are strongly correlated to the Olympic events. Here it is very clear that the Olympic Games are definitely not caused because of the coats, so there is no causation here.
The correlation coefficient is a statistical measure of how strong a relationship exists between two variables' relative movements, with the range of values -1 to 1. A positive correlation corresponds to a directly proportional relationship and when it reaches the value of 1 it shows a perfect positive correlation. A negative correlation denotes an inversely proportionate relationship, and when it reaches the value -1, it is said to be a perfect negative correlation. A correlation of 0.0 indicates that two variables do not have a linear correlation. Pearson and Spearman are the two most prominent and well-known correlation coefficients. The main distinction between the two coefficients is that Pearson deals with linear relationships between two variables, whereas Spearman deals with monotonic relationships as well.
Another distinction is that Pearson uses raw data values for variables, while Spearman uses rank-ordered variables. A monotonic relationship is a relationship whereas the value of one variable increases, the value of another variable also increases or decreases, but not exactly at a constant rate. The rate of increase/decrease in a linear relationship is constant. Following the graphing of pairwise correlations in the dataset (as shown in Figure~\ref{fig:pairplot}), it was discovered that some variables are linearly related, such as DSEP and SE (Figure~\ref{fig:DESP_SE}), while others are monotonically related, such as SE and SP (Figure~\ref{fig:SP_SE}). 

As a result, the spearman coefficient was used to calculate the correlation matrix between the dataset variables. 



The dython library~\cite{dython} (set of data analysis tools in python) was used to compute the correlation matrix.
Since the data includes the variable CLASS which is categorical, the module nominal.associations~\cite{nominal} was used to compute the correlation of association of the features in the data set with categorical and continuous features. 
'clustering=True' causes the computed associations to be sorted into groups by similar correlations. There is a p-value associated with each association, which indicates the chance that the null hypothesis is true. P-value is the measure of the probability that an observed difference could have occurred just by random chance. A p-value determines whether there is evidence to reject a null hypothesis. The greater the difference between two observed values, the less likely it is that the difference is due to random chance, and this is reflected by a lower p-value~\cite{pvalue}. A p-value of 0.05, for example, indicates that there is only a 5 percent possibility that the sample results happened by chance. This means that the outcome is 95 percent guaranteed. In order to uncover patterns and connections between variables as well as between variables and class, the correlation matrix was carefully examined. For a better comprehension of the data, several high positive and negative values, as well as the null values of correlation, were highlighted and analysed. Once a 95\% confidence level was employed, the p-value was used to determine the linear relationship between two variables by comparing the p-value to the significance level with alpha = 0.05. Thus, there is a significant relationship between the variables if p-value is less than or equal 0.05. 

\section{Results}
This Section shows a description of the data collected during the simulations.
\subsection{Features Statistics}
The input parameters in the simulation scenario are variably chosen and have an impact on the physics and duration of the manoeuvres. Because the driver always wants to overtake, his decision will never be influenced by any factor other than the simulation's outcome. However, in the non-simulated real world, one can speculate what circumstances create these results. Following that, some important or anomalous data from each characteristic are studied and interpreted. Reasons for these outcomes are studied given the simulator's characteristics and limitations. A last step consists of speculations on real-world events. 

\input{tables/static_features}
\input{tables/dynamic_features}

Tables~\ref{tab:static_res} and~\ref{tab:dynamic_res} show the measurement results for the static and dynamic features, respectively.
The abbreviations Min and Max represent, respectively, the minimum and maximum value of each feature, and the symbol $\sigma$ represents the standard deviation.
Also in the second table, the abbreviations af and rf represent the absolute and relative frequency, respectively. 


The static features correspond to the total of the observed values according to the different features DN, HL, TE, TP and NV. It can be noted that the data is unbalanced. 
Balancing the data provides the same amount of information for each feature, allowing us to forecast each class more accurately. Now, in this synthetic generated dataset, one can see that there is a significant difference between daytime and nighttime scenes (94 versus 182), as well as scenes with and without the horizon line (250 versus 26). 
It is currently feasible to compare the recorded values for characteristics that calculate the lowest, maximum, and average. When there is no attempt to overtake, the waiting time, which indicates the expected time to overtake, has a high mean value (8.00). It assumes this high value since, for these cases, the vehicle waits until the end of the simulation to overtake.  That is, the value is the same as the simulation's duration. In scenarios where the overtaking is successful and legal, the ego vehicle's driver waits an average of 0.44 seconds before initiating the manoeuvre. In a real-world circumstance, the reverse would be expected: a longer time spent perceiving the surrounding components prior to the manoeuvre would indicate their safety. The overtaking time feature is also quite intuitive.
It is not possible to compute how long it takes in cases when there is no overtaking, hence the number is set to 0. Except for this class, the smallest number relates to an unsuccessful overtaking with collision (2.80). Because the manoeuvre ends when both vehicles collide and the ego vehicle does not return to the original lane, this number is also expected. For unsuccessful overtakes without crashes, the highest rating, 11.16, is recorded. These situations occur when, for some reason, the car does not return to the original lane and instead stays in the left lane until the simulation is completed, which takes longer.

When it comes to dynamic features, for cases of unsuccessful overtaking with collisions, the average value of the ego vehicle speed is high (87.75). When reported for unsuccessful overtaking without crashes, the value is much lower (75.50). This fact may indicate that the overtaking vehicle's speed may have an impact on the occurrence of collisions. On the other hand, the value for the Successful (illegal) class is also high (84.32), so it is not possible to distinguish a successful overtaking from an unsuccessful one by the speed of the ego vehicle alone. When there is no attempt to overtake, the average ego vehicle speed is recorded at its lowest (55.10). It can be assumed that in real-world settings, when a vehicle's speed is low, it will not attempt to pass another vehicle because the latter will most probably drive at a higher speed.

In turn, the average value of the overtaken vehicle speed measurements is consistent, ranging between 57 and 64. The lowest recorded values relate to no overtaking attempt (57.03) and successful and legal overtaking (57.38). The highest value is assigned to the class of unsuccessful overtaking with collisions (64.01). Because they are in such a tiny range, it is clear that the feature Speed of the preceding car does not indicate the class by itself.
The speed difference is an important feature to consider when studying the classes.
The highest reported value corresponds to unsuccessful overtaking situations involving crashes (23.74). Because this value is so near to the values associated with successful overtaking (20.55 and 20.60), it's assumed that the speed difference feature isn't a factor in determining whether an overtaking attempt is successful or not. The value achieved in circumstances where there is no overtaking attempt, on the other hand, is much lower than the others (4.07). This could imply that in situations where the speed differential between the ego vehicle and the other vehicle is insignificant, the driver does not feel the need to overtake.

Distance between C and P is another important factor in feature analysis. It can be expected that the greater the safety distance between vehicles, the more likely that overtaking will be successful. The distances of successful and legal overtaking events have an average value of 55.77. This score is substantially higher than the others, implying that it could be a deciding factor in whether an overtaking is lawful. The lowest rating denotes an unsuccessful overtaking attempt that resulted in a collision (31.50). As a result, it is reasonable to predict that collisions are likely at low safety distances. The high value of 43.78 in non-overtaking scenarios could also indicate that if a car keeps a large distance from the automobile ahead, it has no desire to overtake and will always follow behind it.

The values of the lane occupancy rate to the left of the vehicle that wants to overtake are very similar to each other. The highest and most discordant value is 31.54, which corresponds to no overtaking situations. The lane occupancy does not influence the driver's decision to overtake in the simulation scenario, but it can be a significant effect in real-world settings. Because the lane is more constrained, the driver may be afraid of the manoeuvre due to the risk of colliding.

The observed weather conditions appear to have an impact on class selection. In terms of precipitation, all values are similar, with the exception of the smallest value (33.67), which corresponds to circumstances where there is no overtaking. In the actual world, the driver may be concerned about his overtaking manoeuvre in adverse weather conditions. Because the simulation does not bring into question the driver's judgement to overtake, the data does not support the statement. The same is true for the wind values. The fog values, on the other hand, are distinct.
Cases with successful and lawful overtaking are assigned a lower value of 3.90, whereas cases without overtaking are assigned a lower value of 58.93. These values are due to the fact that fog represents a decrease in visibility which, when overtaking, leads the driver to violate traffic rules, and overtaking is considered illegal. It is expected that in real-world conditions with high fog content, the driver will choose not to overtake for this reason.

\begin{table*}[htbp]
\caption{Information about frame 70 of simulation 1 extracted from dataset.}\label{tab:dataset_frame70}
\centering
\resizebox{\textwidth}{!}{%
\begin{tabular}{@{}|c|c|c|c|c|c|c|c|c|c|c|c|c|c|c|c|c|c|c|c|c|c|@{}}
\toprule
\rowcolor{blue} 
S & F & TS & IDego & Dim & L & V & D & A & MV & RT & LT & LW & LWR & LWL & C & Prec & Fog & Wind & DN & HL & OV \\ \midrule
1 & 70 & 3.5 & 488 & 
\begin{tabular}[c]{@{}c@{}} (488,6.27,2.39,2.1); \\ (489,5.37,1.80,1.57) \\ (490,4.86,2.03,1.65) \end{tabular}
 & \begin{tabular}[c]{@{}c@{}} (488,351.35,251.58,-7); \\ (489,355.82,251.68,-7) \\ (490,318.8,251.61,-7) \end{tabular} & \begin{tabular}[c]{@{}c@{}} (488,66); \\ (489,79) \\ (490,113) \end{tabular} & \begin{tabular}[c]{@{}c@{}} (488,-0.2,-3.62); \\ (489,0.35,-0.43) \\ (490,0.1,0.23) \end{tabular} & \begin{tabular}[c]{@{}c@{}} (488,76.28); \\ (489,40.06) \\ (490, 0.55) \end{tabular} & 90 & Solid & Broken & 3.5 & 0.5 & 3.5 & 489 & 60 & 60 & 60 & Night & No & 0  \\ \bottomrule
\end{tabular}%
}
\end{table*}

\subsection{Data Visualization}
The data collected during each simulation is composed of 22 variables, clearly detailed in Section~\ref{chap:collected_data}. 
Table~\ref{tab:dataset_frame70} shows the data recorded in frame 70 (F) of the first simulation (S) at 3.5 seconds (TS). Three vehicles with ids 488, 489 and 490 are involved, where 488 is the id of the ego vehicle (IDego). These numbers can be checked in the columns Dim, L, V, D and A. Considering only the ego vehicle, its dimension is then 6.27m long, 2.39m wide and 2.1m high (D). It is located in lane -7 having its x and y coordinates equal to 351.35 and 251.58 values, respectively (L). It has a speed of 66 km/h (V) and an acceleration of 76.28 m/s$^2$ (A). The direction of the wheels registers -0.2 on the x-axis and -3.62 on the y-axis (D). In this frame, lane -7 indicates a maximum speed of 90 km/h (MV). The lane to the right of the ego vehicle has the solid type (RT) and to the left the broken type (LT). Considering lane width, the lane in which the ego vehicle is travelling has 3.5 meters wide (LW), the lane to its right has 0.5m (LWR) and the lane to its left has 3.5m (LWL). A collision was recorded in this frame between the ego vehicle and the vehicle with ID 489 (C). As for the weather conditions, the percentage of precipitation, fog and wind registered the same value of 60\% (Prec, Fog, Wind). It is nighttime (DN) and there is no horizon line (HL). Finally, the OV column shows a 0 value, which means that in this frame there was no overtake attempt.

\begin{figure*}[hbtp]
\centering
    \includegraphics[width=\linewidth]{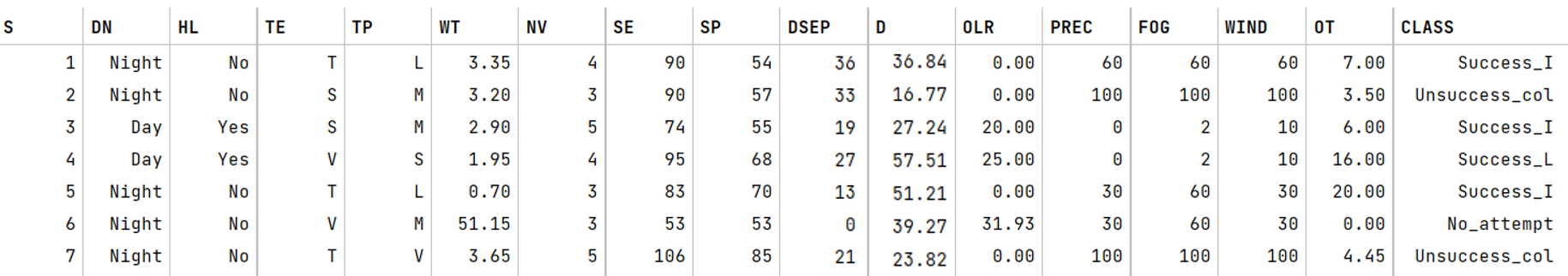}
    \caption{Features Dataset Portion.}
    \label{fig:dataset}
\end{figure*}

The dataset corresponding to the static and dynamic features collected in each simulation was used (Figure~\ref{fig:dataset} ). 



The 'CLASS' column shows the classifications detailed in Section~\ref{chap:classification}. The abbreviation Success\_L corresponds to legal successful overtaking, Success\_I corresponds to illegal successful overtaking, Unsuccess\_col corresponds to unsuccessful overtaking with collisions, Unsuccess\_ncol corresponds to unsuccessful overtaking without collisions, and No\_attempt corresponds to cases where there is no attempt to overtake. In order to standardize the data, the categorical values DN, HL, TP, and TE, are mapped to numerical values. For the DN (Day/Night) feature, the code assigns numerical values so that Day maps to 0 and Night maps to 1. The same logic was followed for the HL (Horizon Line) variable, where Yes maps to 1 and No maps to 0. For the values of the TP and TE columns, vehicles S (small), M (medium), L (large), V (vans), T (truck), MC (motorcycle) and B (bicycle) map respectively the values from 0 to 6. 

\begin{figure}[H]
\centering
\begin{subfigure}[t]{.5\textwidth}
    \centering
    \captionsetup{justification=centering}
    \includegraphics[width=0.95\linewidth]{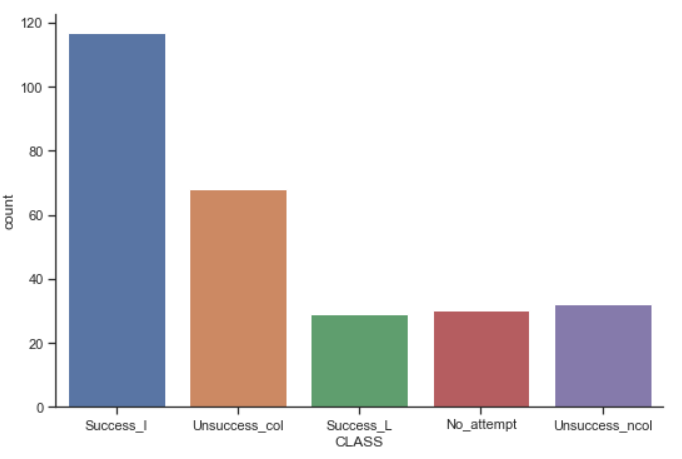}  
\end{subfigure}
\begin{subfigure}[t]{.3\textwidth}
    \centering
    \captionsetup{justification=centering}
    \includegraphics[width=0.95\linewidth]{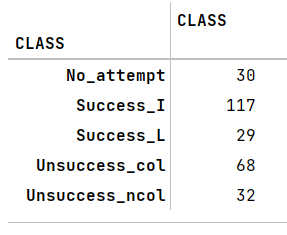}  
\end{subfigure}

\caption{Frequency of classes.}
\label{fig:freq_class}
\end{figure}

To begin data visualization, the histogram and the frequencies of each class are displayed as shown in Figure~\ref{fig:freq_class}.

\begin{figure*}[htbp]
\centering
    \includegraphics[width=\linewidth]{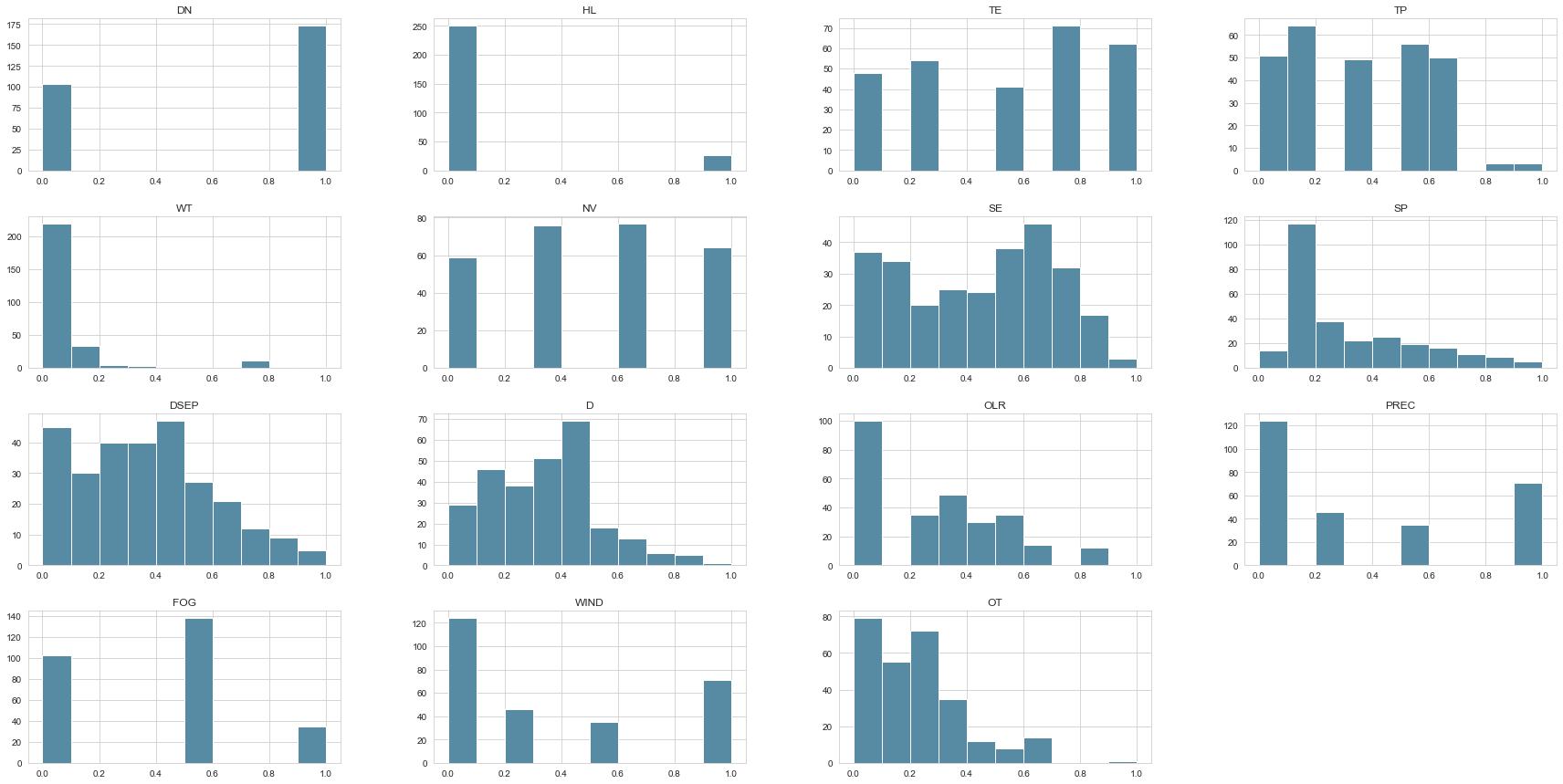}
    \caption{Histogram for each feature.}
    \label{fig:histograms_all}
\end{figure*}

Out of 276 simulations, 30 are classified as No\_attempt, 117 as Success\_I, 29 as Success\_L, 68 as Unsucess\_col and 32 as Unsucess\_ncol. It can be seen that the class with the highest frequency is Success\_I and the one with the lowest frequency is Success\_L.
The total numbers of occurrences for each class would not be as dissimilar if some overruns considered Success\_I were considered Success\_L. This might be because many factors have to happen for the manoeuvre to be considered legal. A histogram for each feature, illustrated in Figure~\ref{fig:histograms_all}, was created to better understand how the data is distributed.

The histogram corresponding to the WT and OT features represents a right-skewed distribution. The data distribution suggests that high values occur with a low frequency. In other words, in the case of the WT variable, the majority of the cases represented in the dataset have a very short time to overtake. The histogram in the OT feature, on the other hand, demonstrates that the overall time of the overtaking manoeuvre is quite low in the majority of situations. This is because the OT value registers a low value in 47 percent of circumstances ((30+68+32)/276*100): in unsuccessful overtaking and when there is no desire to overtake, the value is 0.
The histogram corresponding to the NV (number of vehicles) feature represents a normal distribution. This means that points on one side of the average are as likely to occur as on the other side of the average. About half of the cases have 4 or 5 vehicles present, and another half have 3 or 6 vehicles, the remaining features have a random distribution. In a random distribution histogram, it can be the case that different data properties were combined. In the histogram corresponding to the OLR feature, it is possible to verify that in 36\% (100/276*100) of the cases the average occupancy rate during the simulation is between 0\% and 5\%. In the histogram corresponding to the SP feature (current speed of ego vehicle), about 43\% (120/276*100) of the cases, the ego vehicle drives at a speed between 50 and 55 km/h.
As seen in the histograms of the DN and HL features, there are many more simulations in nighttime scenario than daytime scenario (about 66\% versus 34\%) and many more simulations without the presence of horizon line than with the presence (about 91\% versus 9\%). This is owing to the fact that the horizon line is only present in one of the weather situations considered in the simulations, ClearSunset. In terms of nighttime circumstances, they account for 5 of the 9 presets used, or around 56\%. When the scenarios in each simulation are picked at random, the chances of the features with the biggest number of instances being represented are higher. 

\begin{figure}[H]
    \centering
    \includegraphics[width=0.8\linewidth]{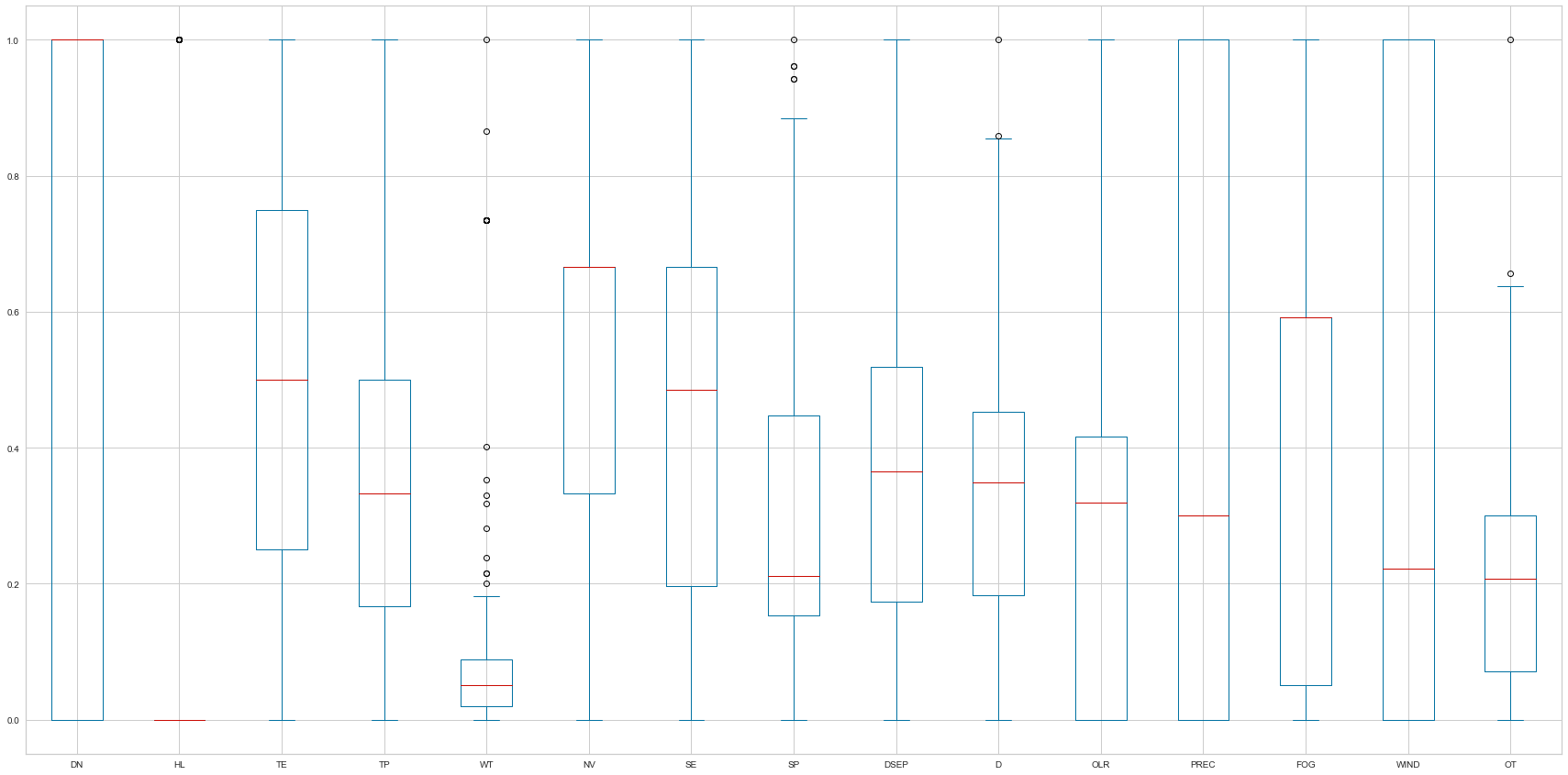}  
    \caption{Box plot diagram for each feature.}
    \label{fig:box-plot}
\end{figure}

Another way of visualizing and interpreting the data was through a box-plot. Figure~\ref{fig:box-plot} shows the box plot diagram for each feature.

It can be observed that, for example, the values of waiting time (WT) are not dispersed, unlike the variables PREC or WIND.  
In the case of feature OT and SP, the median is closer to the bottom quartile. When this happens, then the distribution is not symmetric (normally distributed) but positively skewed (skewed right). The mean is greater than the median, so the data constitute higher frequency of high valued scores.
Many outliers are found in features WT, SP, D and OT. These data should be discarded when they are known to have been entered/measured incorrectly or when they affect assumptions or create significant associations.

\begin{figure}[H]
\centering
\begin{subfigure}[t]{.45\textwidth}
    \centering
    \captionsetup{justification=centering}
    \includegraphics[width=\linewidth]{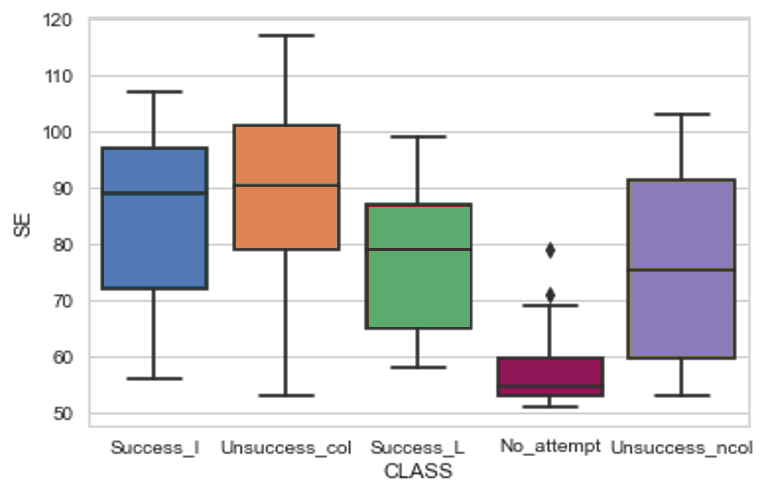}  
    \caption{Box plot.}
    \label{fig:bp_SE}
\end{subfigure}
\begin{subfigure}[t]{.45\textwidth}
    \centering
    \captionsetup{justification=centering}
    \includegraphics[width=\linewidth]{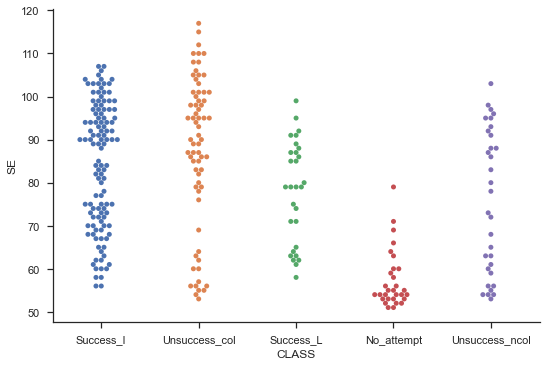}  
    \caption{Swarm plot.}
    \label{fig:sp_SE}
\end{subfigure}
\caption{Box and swarm plots for SE feature grouped by classes.}
\end{figure}

A swarm plot is another way of plotting the distribution of an attribute or the joint distribution of a couple of attributes. The box plot and swarm plot for the SE feature grouped by classes are compared (Figures~\ref{fig:bp_SE} and~\ref{fig:sp_SE}, respectively). 


It can be seen that the swarm plot supports the box plot by highlighting the scattering and clustering zones of the points. It is interesting to see that the box plot of the No\_attempt class is comparatively short. This suggests that, in general, vehicles that do not attempt to overtake have a speed between 55 and 60km/h.

\begin{figure*}[htbp]
    \centering
    \includegraphics[width=0.7\linewidth]{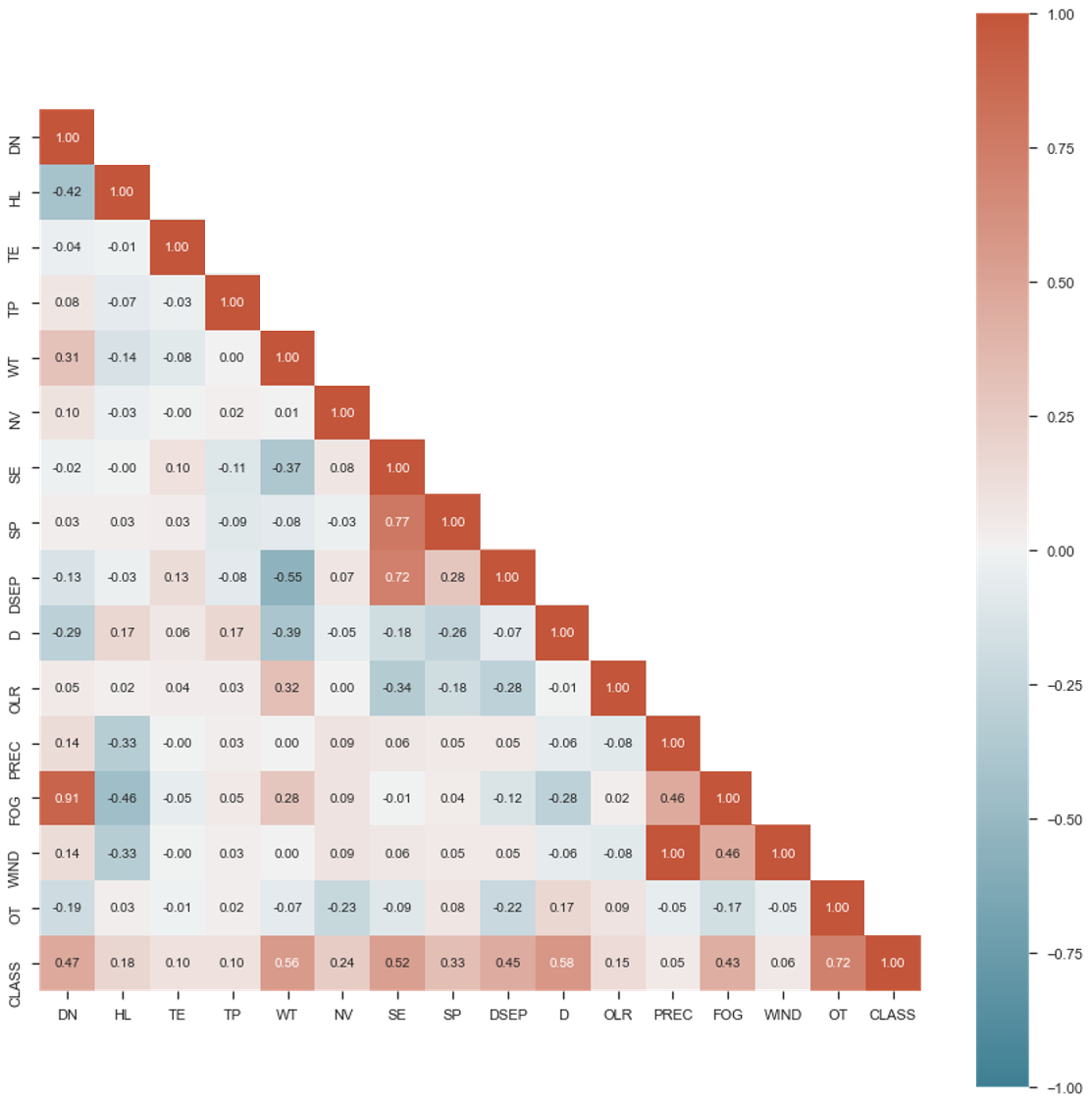} 
    \caption{Correlation matrix.}
    \label{fig:correlated}
\end{figure*}

\begin{figure*}[htbp]
    \centering
    \includegraphics[width=0.9\linewidth]{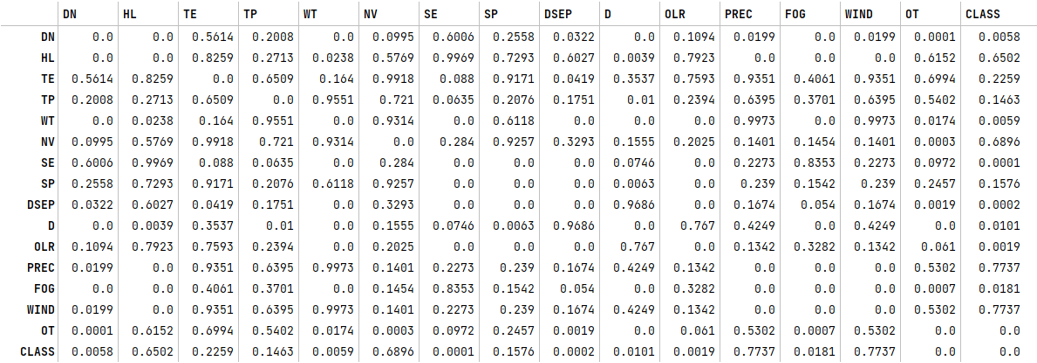}  
    \caption{P-values.}
    \label{fig:pvalues}
\end{figure*}

The correlation matrix was then analysed to uncover relationships among variables and between variables and class. Figure~\ref{fig:correlated} shows the matrix, whereas Figure~\ref{fig:pvalues} shows the corresponding p-values, which are evaluated alongside the matrix. The null hypothesis in this case is a statement that there is no relation between the two variables being compared. Because diagonal elements indicate each variable's association with itself, they will always equal 1.

The correlation between the variables WIND and PREC has a value of 1 indicating a perfect positive correlation. The corresponding p-value of 0 indicates that the null hypothesis (there is no significant relationship between the two variables) can be rejected. 
This occurs because each precipitation value correlates to the same wind value in the presets specified for the meteorological circumstances of the simulations.
Another example of positive correlation is the 0.91 correlation between the FOG and DN variables, with a p-value of 0. It means that higher fog intensity is significantly linked to nighttime scenarios. 
Since it is expected that the simulated data would accurately reflect reality, it is important to discuss potential relationships that are discovered. As such, the two cases presented above are distinct in that the strong relationship of the first example is expected as opposed to the strong relationship of the second case. In the non-simulated real world, strong wind is often observed on days of heavy rain. In the second case, it is believed that in the real world this relationship would not be so evident since clear nights without fog can often be found.

The SE and DSEP variables, on the other hand, have a positive correlation of 0.72. Then, because the p-value is 0, it can be concluded with a high level of confidence that for high ego vehicle speeds, a large difference in speeds between the ego vehicle and the overtaken vehicle is observed. As for the negative correlations, there is no perfectly negative one. With a value of -0.55, the correlation between the DSEP and WT variables is the most significant. The relevance of this relation is confirmed by the p-value of 0. In this case, it turns out that the waiting time to overtake is greater when two cars are travelling at similar speeds, i.e., the speed difference is low. In a real-life scenario, this statement makes sense since for an overtaking to occur and the wait time to be reduced, the vehicle must accelerate and move closer to the car in front, increasing the DSEP value. For near-zero correlation values, such as the case of the 0.075 value for the NV and SE variables, indicates that they're basically not correlated. The p-value of 0.284 confirms this value, indicating that there is about 28 percent possibility that the results were obtained by chance. As a result, the number of vehicles in the simulation and the ego vehicle speed have almost no relationship. Analyzing this example might be interesting. In the simulated situation, the number of cars in the space under consideration for simulation has no impact on the ego vehicle's speed. In a real-world scenario, it would be reasonable to assume that the ego vehicle speed would be lower with a heavier traffic flow. However, when comparing only the two variables - number of vehicles and ego vehicle speed - there is no information about the speeds of the other vehicles. Assuming that all vehicles are moving at constant and similar speeds, the speed of the ego vehicle would not need to be changed. This value of correlation could be increased if the simulation window was reduced and the number of possible cars was increased.

\balance
\section{Conclusions}
The challenges of autonomous driving resolve around accurately perceiving the environment surrounding the autonomous car, understanding and distinguishing the various elements that constitute the scene, but cars also require mechanisms that support decision-making. One crucial aspect of a good decision-making is the data and knowledge representation of the world with all its objects and their interactions. Most of the literature up to date concentrates on sensor data perception such as semantic segmentation, labeling, object detection and object uncertainty. The quality of these types of tasks is still low, specially under adverse conditions (brightness, darkness, fog, among others). Vehicles in the market have yet to reach level 3, given that a combination of perception, planning, decision making, and control is not very mature yet. The main contributions of this work are: (1) we provide a vast literature review gathering the main studies in the area of autonomous driving, including the status of perception mechanisms and datasets; (2) we provide a thorough discussion about important features to be taken into account for overtaking manoeuvre; (3) we provide a synthetic dataset collected through simulation which takes into account several important factors to be considered for decision-making during overtaking. Finally, we describe our synthetic dataset and provide feedback on its main characteristics. 


\section*{Acknowledgment}
This work is supported by European Structural and Investment Funds in the FEDER component, through the Operational Competitiveness and Internationalization Programme (COMPETE 2020) [Project \# 047264 -- THEIA; Funding Reference: POCI-01-0247-FEDER-047264].

%
% ...


\bibliographystyle{asmejour}

\bibliography{refs}

\clearpage
\onecolumn
\appendix
\include{appendix}

\end{document}

%% file: tables/variables.tex
\begin{table}[H]
\centering
\caption{Stored variables in each simulation.\\ 
Legend: idv represents the vehicle id and id\_lane represents the id of the lane the vehicle is in.}
\label{tab:data_collected}
\resizebox{\columnwidth}{!}{%
\begin{tabular}{@{}|l|l|l|@{}}
\toprule
\rowcolor{blue} 
Name & Description & Unit of Measurement \\ \midrule
S & Number of simulation & int \\ \midrule
F & Frame & int \\ \midrule
TS & Timestamp & seconds \\ \midrule
IDego & Id of ego vehicle & int \\ \midrule
Dim & Vehicle Dimension & (idv, x, y, z) \\ \midrule
L & Vehicle location  & (idv, x, y, id\_lane) \\ \midrule
V & Vehicle velocity  & (idv, km/h) \\ \midrule
D & Vehicle direction  & (idv, x, y) \\ \midrule
A & Vehicle acceleration & (idv, m/s$^2$) \\ \midrule
MV & \begin{tabular}[c]{@{}c@{}} Max velocity corresponding to the \\ ego vehicle lane \end{tabular} & km/h \\ \midrule
RT & Type of right line of the ego vehicle & "Solid", "Broken", etc \\ \midrule
LT & Type of left line of the ego vehicle & "Solid", "Broken", etc \\ \midrule
LW & Width of ego vehicle current lane & m \\ \midrule
LWR & Width of right lane & (m, id\_lane) \\ \midrule
LWL & Width of left lane & (m, id\_lane) \\ \midrule
C & \begin{tabular}[c]{@{}c@{}} Collision between ego vehicle \\ and other object \end{tabular}  & idv \\ \midrule
Prec & Precipitation  & \% \\ \midrule
Fog & Fog  & \% \\ \midrule
Wind & Wind & \%  \\ \midrule
DN & Day or night & Day or Night \\ \midrule
HL & Horizon line & Yes or No \\ \midrule
OV & \begin{tabular}[c]{@{}c@{}} Time when the overtake occurs \\ (1 if true, 0 if false) \end{tabular}  & 0 or 1 \\ \bottomrule

\end{tabular}%
}
\end{table}

%% file: tables/static_features.tex
\begin{table*}[htbp]
\caption{Statistics of Static Features.}
\label{tab:static_res}
\centering
\resizebox{\textwidth}{!}{%
\begin{tabular}{|c|cccccccccccccc|c}
\hline
\rowcolor{blue}
STATIC   FEATURES                           & \multicolumn{14}{c|}{\cellcolor{blue}STATICS}                                                                                                                                                                                                                                                                                                                                                                                                                                                                                                                                                                                                                                                & \multicolumn{1}{c|}{\cellcolor{blue}OUTCOME}              \\ \hline
                                            & \multicolumn{7}{c|}{\cellcolor[HTML]{E2F0D9}Day}                                                                                                                                                                                                                                                                                                            & \multicolumn{7}{c|}{\cellcolor[HTML]{E2F0D9}Night}                                                                                                                                                                                                                                                                                     &                                                                   \\
\hline
\end{tabular}%
}

\end{table*}

%% file: tables/dynamic_features.tex
\begin{table*}[htbp]
\centering
\caption{Statistics of Dynamic Features.}\label{tab:dynamic_res}
\begin{tabular}{@{}|cc|cccc|c|@{}}

\hline
\multicolumn{2}{|c|}{\cellcolor{blue}} &
  \multicolumn{4}{c|}{\cellcolor{blue}{STATISTICS}} &
  \cellcolor{blue} \\ \cline{3-6}

\multicolumn{2}{|c|}{\multirow{-2}{*}{\cellcolor{blue}DYNAMIC FEATURES}} &
  \multicolumn{1}{c|}{\cellcolor[HTML]{E2F0D9}Min} &
  \multicolumn{1}{c|}{\cellcolor[HTML]{E2F0D9}Mean} &
  \multicolumn{1}{c|}{\cellcolor[HTML]{E2F0D9} Max } &
  \multicolumn{1}{c|}{ \cellcolor[HTML]{E2F0D9} $\sigma$ } &
  \multirow{-2}{*}{\cellcolor{blue}{OUTCOME}} \\ \hline

\multicolumn{2}{|c|}{}                                             & \multicolumn{1}{c|}{58.00} & \multicolumn{1}{c|}{77.93} & \multicolumn{1}{c|}{99.00}  & 11.89 & Successful (legal)\\ \cline{3-7} 
\multicolumn{2}{|c|}{}                                             & \multicolumn{1}{c|}{56.00} & \multicolumn{1}{c|}{84.32} & \multicolumn{1}{c|}{107.00} & 14.44 & Successful (illegal)\\ \cline{3-7} 
\multicolumn{2}{|c|}{}                                             & \multicolumn{1}{c|}{53.00} & \multicolumn{1}{c|}{75.50} & \multicolumn{1}{c|}{103.00} & 16.71 & Non-successful (no collision)\\ \cline{3-7} 
\multicolumn{2}{|c|}{}                                             & \multicolumn{1}{c|}{53.00} & \multicolumn{1}{c|}{87.75} & \multicolumn{1}{c|}{117.00} & 17.66 & Non-successful (collision)\\ \cline{3-7} 
\multicolumn{2}{|c|}{\multirow{-5}{*}{Current Speed}}              & \multicolumn{1}{c|}{53.00} & \multicolumn{1}{c|}{55.10} & \multicolumn{1}{c|}{65.00}  & 2.64  & No attempt to overtake \\ \hline
\multicolumn{2}{|c|}{}                                             & \multicolumn{1}{c|}{50.00} & \multicolumn{1}{c|}{57.38} & \multicolumn{1}{c|}{78.00}  & 7.49  & Successful (legal)\\
\hline
  
\end{tabular}%
\end{table*}

%% file: appendix.tex
\section{Currently systems implemented in autonomous vehicles}\label{appendixA}
\input{tables/current_systems}

\newpage

\section{Comparison of main datasets' features}\label{appendixB}
\input{tables/comparison_datasets}

%% file: tables/current_systems.tex
\newcolumntype{A}{>{\centering\arraybackslash}m{0.12\textwidth}}
\newcolumntype{B}{>{\centering\arraybackslash}m{0.52\textwidth}}
\newcolumntype{C}{>{\centering\arraybackslash}m{0.05\textwidth}}
\newcolumntype{D}{>{\centering\arraybackslash}m{0.08\textwidth}}
\newcolumntype{E}{>{\centering\arraybackslash}m{0.06\textwidth}}

\begin{longtable}{|A|B|C|D|E|}
\caption{Currently systems implemented in autonomous vehicles~\cite{tesla_systems}~\cite{bosch_systems}~\cite{bmw_2021}. \\ Comfort and Safety are represented on a scale of 1 to 5:1 is (--), 2 is (-), 3 is (+), 4 and 5 are (++).}
\label{tab:automated_systems} 
\\
\hline
\rowcolor{blue} 
System & Description & SAE Level & Comfort & Safety \\ \midrule
\endfirsthead
\multicolumn{5}{c}%
{{\bfseries Table \thetable\ continued from previous page}} \\ \midrule
\rowcolor{blue} 
System & Description & SAE Level & Comfort & Safety \\ \midrule
\endhead
Hill Start Assist &  Makes the rear axle secures the vehicle for a  short time when the driver disengages the brake.   & 0 & ++ & -      \\ \midrule
Road Sign Recognition & Recognizes traffic signs, road lines, and other road elements, adapts driving while displaying  all the information directly on the car's screen. & 0         & ++      & +      \\ \midrule
Emergency Brake Assist &  Brings the vehicle to a complete stop before it  results in a crash with an obstruction, a person  or another vehicle. & 0         & -      & ++      \\ \midrule

Steering and Lane Guidance Assist & Can help the driver to drive the vehicle in the middle of the lane and therefore relieve him from steering. & 1         & ++      & ++      \\ \midrule

Cruise Control & The system constantly maintains the driver chosen speed. & 1         & ++      & --      \\ \midrule

Adaptive Cruise Control & Adjusts the vehicle speed to match the traffic flow and/or maintain a pre-set distance to the vehicle in front. & 1         & ++      & --      \\ \midrule

Lane-change Warning & Alert the driver if there is any danger of overtaking monitoring the area beside and behind the car, and covering the notorious blind spot. & 2         & +      & ++      \\ \midrule

Parking Assist & Support the driver when manoeuvring or parking and help protect him from parking damage. & 3         & ++      & -      \\ \midrule

Driving Pilot & The vehicle should be able to take over driving completely so the driver can take their hands off the wheel for a certain amount of time. However, the driver must always be capable of resuming driving within a certain amount of time when requested to do so. & 3         & ++      & +      \\ \midrule

Valet Parking & Some companies have developed some systems like “Automated valet parking” from Bosch or “Intelligent Park Pilot” from Mercedes~\cite{mercedes_systems}. The system takes the hassle out of searching for a parking spot but also handles the actual task of parking the car in the parking garage. & 4         & ++      & +     \\ \bottomrule

\end{longtable}

%% file: tables/comparison_datasets.tex
\begin{table}[htp]
\centering
\caption{Comparison of main datasets' features. Adapted from~\cite{9156412} and~\cite{Laflamme2019Oct}. \\ Legend: (-) indicates that no information is provided. X means No and \checkmark means Yes.\label{tab:datasets_comp}}

\resizebox{\textwidth}{!}{%
\begin{tabular}{@{}|c|c|c|c|c|c|c|c|c|c|c|c|c|c|c|c|@{}}
\toprule
\rowcolor{blue} 
Dataset                              & Year & \begin{tabular}[c]{@{}c@{}}\# \\ Scenes\end{tabular} & \begin{tabular}[c]{@{}c@{}}Size\\ (h)\end{tabular} & \begin{tabular}[c]{@{}c@{}}\# RGB \\ Images\end{tabular} & \begin{tabular}[c]{@{}c@{}}\# pc \\ LiDaR\end{tabular} & \begin{tabular}[c]{@{}c@{}}\# pc\\ Radar\end{tabular} & \begin{tabular}[c]{@{}c@{}}\# ann.\\ frames\end{tabular} & \begin{tabular}[c]{@{}c@{}}\ 3D \\ boxes\end{tabular} & \begin{tabular}[c]{@{}c@{}}\# Map\\ Layers\end{tabular} & \begin{tabular}[c]{@{}c@{}}\# \\ Classes\end{tabular} & 
Night & Rain & Snow & Dawn 
& Locations 
\\ \midrule
\cellcolor{blue} Kitti        & 2012 & 22                                                   & 1.5                                                & 15k                                                      & 15k                                                    & 0                                                     & 15k                                                      & \checkmark                                                   & 0                                                       & 8                                                     & X    & X   & X   & X   & Karlsruhe        \\ \midrule
\cellcolor{blue}Cityscapes   & 2016 & -                                                    & -                                                  & 25k                                                      & 0                                                      & 0                                                     & 25k                                                      & X                                                     & 0                                                       & 30                                                    & X   & X   & X   & X   & 50 cities         \\ \midrule
\cellcolor{blue}CamVid       & 2008 & 4                                                    & 0.4                                                & 18k                                                      & 0                                                      & 0                                                     & 700                                                      & X                                                      & 0                                                       & 32                                                    & X    & X   & X   & X   & Cambridge         \\ \midrule
\cellcolor{blue}Vistas       & 2017 & -                                                    & -                                                  & 25k                                                      & 0                                                     & 0                                                     & 25k                                                      & X                                                      & 0                                                       & 152                                                   & \checkmark   & \checkmark  & \checkmark  & \checkmark  & 6 continents      \\ \midrule
\cellcolor{blue}Waymo Open   & 2019 & 1k                                                   & 5.5                                                & 1M                                                       & 200k                                                   & 0                                                     & 200k                                                     & \checkmark                                                    & 0                                                       & 4                                                     & \checkmark   & \checkmark  & X   & \checkmark  & USA               \\ \midrule
\cellcolor{blue}Lyft Level 5 & 2019 & 366                                                  & 2.5                                                & 323k                                                     & 46k                                                    & 0                                                     & 46k                                                      & \checkmark                                                   & 7                                                       & 9                                                     & X    & X   & X   & X   & Palo Alto         \\ \midrule
\cellcolor{blue}nuScenes     & 2019 & 1k                                                   & 5.5                                                & 1.4M                                                     & 400k                                                   & 1.3M                                                  & 40k                                                      & \checkmark                                                   & 11                                                      & 23                                                    & \checkmark   & \checkmark  & X   & \checkmark  & Boston, Singapura \\ \midrule
\cellcolor{blue}ApolloScape  & 2018 & 147k                                                 & 100                                                & 144k                                                     & 0                                                      & 0                                                     & 144k                                                     & \checkmark                                                   & 0                                                       & 8-35                                                  & \checkmark   & \checkmark  & \checkmark  & X   & China             \\ \midrule
\cellcolor{blue}KAIST        & 2018 & -                                                    & -                                                  & 8.9k                                                     & 8.9k                                                   & 0                                                     & 8.9k                                                     & X                                                      & 0                                                       & 3                                                     & \checkmark   & X   & X   & \checkmark  & Seoul             \\ \midrule
\cellcolor{blue}BDD100K      & 2017 & 100k                                                 & 1k                                                 & 100M                                                     & 0                                                      & 0                                                     & 100k                                                     & X                                                      & 0                                                       & 10                                                    & \checkmark   & \checkmark  & \checkmark  & \checkmark  & NY, San Francisco \\ \midrule
\cellcolor{blue}H3D          & 2019 & 160                                                  & 0.77                                               & 83k                                                      & 27k                                                    & 0                                                     & 27k                                                      & \checkmark                                                   & 0                                                       & 8                                                     & X    & X   & X   & X   & San Francisco     \\ \midrule
\cellcolor{blue}PandaSet     & 2020 & 103 & 1k & 48k & 16k & 0 & 8240 & \checkmark & 15 & 28 & \checkmark & \checkmark & \checkmark & X & San Francisco  \\ \midrule
\cellcolor{blue}ACDC         & 2021 & 4                                                  & -                                               & 4k                                                     & 0                                                    & 0                                                    & 4k(?)                                                      & \checkmark                                                   & 0                                                       & 19                                                   & \checkmark    & \checkmark   & \checkmark   & X   & Switzerland     \\ \bottomrule

\end{tabular}%
}

\vspace{1em}
\resizebox{\textwidth}{!}{%
\begin{tabular}{@{}|
>{\columncolor{blue}}c |ccclc|clcll|lllll|@{}}
\toprule
\cellcolor{blue}                          & \multicolumn{5}{c|}{\cellcolor{blue}Sensors}                                                                                                                                             & \multicolumn{5}{c|}{\cellcolor{blue}Traffic}                                                                                               & \multicolumn{5}{c|}{\cellcolor{blue}Annotation}                                                                                                                                     \\ \cmidrule(l){2-16} 
\multirow{-2}{*}{\cellcolor{blue}Dataset} & \multicolumn{1}{l|}{Video} & \multicolumn{1}{l|}{LiDaR} & \multicolumn{1}{l|}{GPS} & \multicolumn{1}{l|}{\begin{tabular}[c]{@{}l@{}}Thermal \\ Camera\end{tabular}} & \multicolumn{1}{l|}{Radar} & \multicolumn{1}{l|}{Urban} & \multicolumn{1}{l|}{Highway} & \multicolumn{1}{l|}{Residential} & \multicolumn{1}{l|}{Rural} & Campus                 & \multicolumn{1}{l|}{Pixel} & \multicolumn{1}{l|}{Instance} & \multicolumn{1}{l|}{\begin{tabular}[c]{@{}l@{}}Point \\ Cloud\end{tabular}} & \multicolumn{1}{l|}{2D} & 3D                     \\ \midrule
Kitti                                             & \multicolumn{1}{c|}{\checkmark}     & \multicolumn{1}{c|}{\checkmark}     & \multicolumn{1}{c|}{\checkmark}   & \multicolumn{1}{l|}{}                                                          &                            & \multicolumn{1}{c|}{\checkmark}     & \multicolumn{1}{c|}{\checkmark}       & \multicolumn{1}{c|}{\checkmark}           & \multicolumn{1}{c|}{\checkmark}     & \multicolumn{1}{c|}{\checkmark} & \multicolumn{1}{l|}{}      & \multicolumn{1}{l|}{}         & \multicolumn{1}{l|}{}                                                       & \multicolumn{1}{c|}{\checkmark}  & \multicolumn{1}{c|}{\checkmark} \\ \midrule
Cityscapes                                        & \multicolumn{1}{c|}{\checkmark}     & \multicolumn{1}{l|}{}      & \multicolumn{1}{c|}{\checkmark}   & \multicolumn{1}{l|}{}                                                          &                            & \multicolumn{1}{c|}{\checkmark}     & \multicolumn{1}{l|}{}        & \multicolumn{1}{l|}{}            & \multicolumn{1}{l|}{}      &                        & \multicolumn{1}{c|}{\checkmark}     & \multicolumn{1}{c|}{\checkmark}        & \multicolumn{1}{l|}{}                                                       & \multicolumn{1}{l|}{}   &                        \\ \midrule
CamVid                                            & \multicolumn{1}{c|}{\checkmark}     & \multicolumn{1}{l|}{}      & \multicolumn{1}{l|}{}    & \multicolumn{1}{l|}{}                                                          &                            & \multicolumn{1}{c|}{\checkmark}     & \multicolumn{1}{l|}{}        & \multicolumn{1}{l|}{}            & \multicolumn{1}{l|}{}      &                        & \multicolumn{1}{c|}{\checkmark}     & \multicolumn{1}{l|}{}         & \multicolumn{1}{l|}{}                                                       & \multicolumn{1}{l|}{}   &                        \\ \midrule
Vistas                                            & \multicolumn{1}{c|}{\checkmark}     & \multicolumn{1}{l|}{}      & \multicolumn{1}{l|}{}    & \multicolumn{1}{l|}{}                                                          &                            & \multicolumn{1}{c|}{\checkmark}     & \multicolumn{1}{c|}{\checkmark}       & \multicolumn{1}{c|}{\checkmark}           & \multicolumn{1}{c|}{\checkmark}     &                        & \multicolumn{1}{c|}{\checkmark}     & \multicolumn{1}{c|}{\checkmark}        & \multicolumn{1}{l|}{}                                                       & \multicolumn{1}{l|}{}   &                        \\ \midrule
Waymo Open                                        & \multicolumn{1}{c|}{\checkmark}     & \multicolumn{1}{c|}{\checkmark}     & \multicolumn{1}{l|}{}    & \multicolumn{1}{l|}{}                                                          &                            & \multicolumn{1}{c|}{\checkmark}     & \multicolumn{1}{c|}{\checkmark}       & \multicolumn{1}{c|}{\checkmark}           & \multicolumn{1}{c|}{\checkmark}     &                        & \multicolumn{1}{l|}{}      & \multicolumn{1}{l|}{}         & \multicolumn{1}{l|}{}                                                       & \multicolumn{1}{c|}{\checkmark}  & \multicolumn{1}{c|}{\checkmark} \\ \midrule
Lyft Level 5                                      & \multicolumn{1}{c|}{\checkmark}     & \multicolumn{1}{c|}{\checkmark}     & \multicolumn{1}{c|}{\checkmark}   & \multicolumn{1}{l|}{}                                                          &                            & \multicolumn{1}{c|}{\checkmark}     & \multicolumn{1}{l|}{}        & \multicolumn{1}{c|}{\checkmark}           & \multicolumn{1}{l|}{}      &                        & \multicolumn{1}{l|}{}      & \multicolumn{1}{l|}{}         & \multicolumn{1}{l|}{}                                                       & \multicolumn{1}{c|}{\checkmark}  & \multicolumn{1}{c|}{\checkmark} \\ \midrule
nuScenes                                          & \multicolumn{1}{c|}{\checkmark}     & \multicolumn{1}{c|}{\checkmark}     & \multicolumn{1}{c|}{\checkmark}   & \multicolumn{1}{l|}{}                                                          & \checkmark                          & \multicolumn{1}{c|}{\checkmark}     & \multicolumn{1}{l|}{}        & \multicolumn{1}{l|}{}            & \multicolumn{1}{l|}{}      &                        & \multicolumn{1}{l|}{}      & \multicolumn{1}{l|}{}         & \multicolumn{1}{l|}{}                                                       & \multicolumn{1}{c|}{\checkmark}  & \multicolumn{1}{c|}{\checkmark} \\ \midrule
ApolloScape                                       & \multicolumn{1}{c|}{\checkmark}     & \multicolumn{1}{c|}{\checkmark}     & \multicolumn{1}{c|}{\checkmark}   & \multicolumn{1}{l|}{}                                                          &                            & \multicolumn{1}{c|}{\checkmark}     & \multicolumn{1}{c|}{\checkmark}       & \multicolumn{1}{c|}{\checkmark}           & \multicolumn{1}{c|}{\checkmark}     &                        & \multicolumn{1}{c|}{\checkmark}     & \multicolumn{1}{c|}{\checkmark}        & \multicolumn{1}{c|}{\checkmark}                                                      & \multicolumn{1}{l|}{}   &                        \\ \midrule
KAIST                                             & \multicolumn{1}{c|}{\checkmark}     & \multicolumn{1}{c|}{\checkmark}     & \multicolumn{1}{c|}{\checkmark}   & \multicolumn{1}{c|}{\checkmark}                                                         &                            & \multicolumn{1}{c|}{\checkmark}     & \multicolumn{1}{l|}{}        & \multicolumn{1}{c|}{\checkmark}           & \multicolumn{1}{l|}{}      & \multicolumn{1}{c|}{\checkmark} & \multicolumn{1}{l|}{}      & \multicolumn{1}{l|}{}         & \multicolumn{1}{l|}{}                                                       & \multicolumn{1}{c|}{\checkmark}  &                        \\ \midrule
BDD100K                                           & \multicolumn{1}{c|}{\checkmark}     & \multicolumn{1}{l|}{}      & \multicolumn{1}{c|}{\checkmark}   & \multicolumn{1}{l|}{}                                                          &                            & \multicolumn{1}{c|}{\checkmark}     & \multicolumn{1}{c|}{\checkmark}       & \multicolumn{1}{c|}{\checkmark}           & \multicolumn{1}{c|}{\checkmark}     &                        & \multicolumn{1}{c|}{\checkmark}     & \multicolumn{1}{c|}{\checkmark}        & \multicolumn{1}{l|}{}                                                       & \multicolumn{1}{l|}{}   &                        \\ \midrule
H3D                                               & \multicolumn{1}{c|}{\checkmark}     & \multicolumn{1}{c|}{\checkmark}     & \multicolumn{1}{c|}{\checkmark}   & \multicolumn{1}{l|}{}                                                          &                            & \multicolumn{1}{c|}{\checkmark}     & \multicolumn{1}{l|}{}        & \multicolumn{1}{l|}{}            & \multicolumn{1}{l|}{}      &                        & \multicolumn{1}{l|}{}      & \multicolumn{1}{l|}{}         & \multicolumn{1}{l|}{}                                                       & \multicolumn{1}{l|}{}   & \multicolumn{1}{c|}{\checkmark} \\ \midrule
PandaSet   & \multicolumn{1}{c|}{\checkmark}     & \multicolumn{1}{c|}{\checkmark}     & \multicolumn{1}{c|}{\checkmark}   & \multicolumn{1}{l|}{}                                                          &                            & \multicolumn{1}{c|}{\checkmark}     & \multicolumn{1}{c|}{\checkmark}       & \multicolumn{1}{c|}{\checkmark}           & \multicolumn{1}{c|}{\checkmark}     &                        & \multicolumn{1}{c|}{\checkmark}     & \multicolumn{1}{c|}{\checkmark}        & \multicolumn{1}{c|}{\checkmark}                                                      & \multicolumn{1}{c|}{\checkmark}   &  \multicolumn{1}{c|}{\checkmark} \\ \midrule
ACDC                                               & \multicolumn{1}{c|}{\checkmark}     & \multicolumn{1}{c|}{}     & \multicolumn{1}{c|}{\checkmark}   & \multicolumn{1}{l|}{}                                                          &                            & \multicolumn{1}{c|}{\checkmark}     & \multicolumn{1}{c|}{\checkmark}        & \multicolumn{1}{l|}{}            & \multicolumn{1}{l|}{\checkmark}      &                        & \multicolumn{1}{c|}{\checkmark}      & \multicolumn{1}{c|}{}         & \multicolumn{1}{l|}{}                                                       & \multicolumn{1}{l|}{}   & \multicolumn{1}{c|}{\checkmark} \\

\bottomrule
\end{tabular}%
}
\end{table}